\RequirePackage{fix-cm}
\documentclass[smallextended]{svjour3}       \smartqed  \usepackage{graphicx}
\usepackage{natbib}
\usepackage{xcolor}

\usepackage[utf8]{inputenc} \usepackage[T1]{fontenc}    \usepackage{hyperref}       \usepackage{url}            \usepackage{booktabs}       

\usepackage{multirow}

\usepackage{geometry}
\usepackage{pdflscape}
\usepackage{rotating}

\usepackage{subcaption}
\usepackage{multirow}

\usepackage{adjustbox}

\usepackage{listings}
\newcommand{\reviseone}[1]{\textcolor{black}{#1}}  
\begin{document}

\title{
Narrative Cartography with Knowledge Graphs
}
\subtitle{}


\author{Gengchen Mai  \and Weiming Huang  \and Ling Cai  \and Rui Zhu  \and Ni Lao        
}

\authorrunning{Gengchen Mai et al.} 

\institute{Gengchen Mai \at
	STKO Lab, Department of Geography, UC Santa Barbara, CA, USA, 93106  \\
	Department of Computer Science, Stanford University, Stanford, CA, USA, 94305 \\
	\email{gengchen\_mai@geog.ucsb.edu}            \\
	\and
	Weiming Hunag$*$ \at
	Department of Physical Geography and Ecosystem Science, Lund University, S-223 62, Lund, Sweden \\
	\email{weiming.huang@nateko.lu.se}  
	\and
	Ling Cai \at
	STKO Lab, Department of Geography, UC Santa Barbara, CA, USA, 93106  \\
	\email{lingcai@ucsb.edu}            \\
	\and
	Rui Zhu \at
	STKO Lab, Department of Geography, UC Santa Barbara, CA, USA, 93106  \\
	\email{ruizhu@ucsb.edu}            \\
	\and
	Ni Lao \\
	\email{noon99@gmail.com }            \\
}

\date{Received: date / Accepted: date}

\maketitle

\begin{abstract}
Narrative cartography is a discipline which studies the interwoven nature of stories and maps. \reviseone{However, conventional geovisualization techniques of narratives often encounter} several prominent challenges, including \reviseone{the data acquisition \& integration challenge and the semantic challenge}. To \reviseone{tackle} these challenges, in this paper, we propose the idea of narrative cartography with knowledge graphs \reviseone{(KGs)}. \reviseone{Firstly, to tackle the data acquisition \& integration challenge, we develop a set of KG-based GeoEnrichment toolboxes to allow users to search and retrieve relevant data from integrated cross-domain knowledge graphs for narrative mapping from within a GISystem. }
With the help of this tool, \reviseone{the retrieved data from KGs are directly materialized in a GIS format which is ready for spatial analysis and mapping}. Two use cases - Magellan's expedition and World War II - are presented to show the effectiveness of this approach. \reviseone{In the meantime,} several limitations are identified from this approach, such as data incompleteness, semantic incompatibility, and the semantic challenge in geovisualization. \reviseone{For the later two limitations,} we propose a modular ontology for narrative cartography, which \reviseone{formalizes} both the map content (\textit{Map Content} Module) and the geovisualization process (\textit{Cartography} Module). We demonstrate that, by representing both the map content and the geovisualization process in KGs (an ontology), we can realize both data reusability and map reproducibility for narrative cartography.

\keywords{Narrative Cartography \and Knowledge Graph \and Semantic Web \and Geospatial Semantics \and Ontology Design Pattern}
\end{abstract} \section{introduction} \label{sec:intro}
Maps, as symbolic representations for the spatial locations of and relationships among elements in space, are widely adopted as an effective instrument for disseminating, and comprehending the spatial dimension (and also temporal dimension in many cases) of various geographies. Such geographies include the ones in the real world, and in fictional worlds (e.g., in fictions, films, and games). Maps thereof have become a regular proxy to better understand the narratives that embodied in the geographies, primarily in terms of \reviseone{the} placement, and \reviseone{the} evolution of the narratives. For example, maps are used to describe the progression of historical events such as migrations, wars, and accidents \citep{caquard2014narrative}. In addition, maps are utilized to systematically study the geography of fictional narratives such as \textit{Atlas of Literature} \citep{bradbury1998atlas}, \textit{Atlas of the European novel, 1800-1900} \citep{moretti1998atlas}, and the research project \textit{A Literary Atlas of Europe}\footnote{\url{http://www.literaturatlas.eu/en/}}.

Narrative cartography \citep{caquard2013cartography,caquard2014narrative,ryan2020narrative} is a discipline which studies the interwoven and intimate relationship between stories and maps. The materialization of such intimate relationship evolves over time for centuries with the development of technologies. 
In old times, mapping processes were mostly accomplished on papers and through labor-intensive creation \citep{varanka2018map}. 
Today, with the rapid development of web mapping technology, online web maps (e.g., Google Maps, Apple Maps, OpenStreetMap, and Baidu Map) have become an indispensable component and one of the major sources of geographic information in our daily lives.

Thanks to the arising of Web 2.0 and the development of web mapping and GPS technologies, an increasing number of user-generated contents, such as social media posts, photos, or even audios, are associated with geotags to depict users' activities in the geographic space \citep{amoroso2010exposed}. Such kinds of volunteered geographic information (VGI) \citep{goodchild2007citizens} provide a geospatial view of personal stories for users. Additionally, the studies on temporal aspects of mapping considerably enhance the capability of maps to be a vivid and perceptible channel to delineate stories that evolve over time; see e.g., \citet{andrienko2010space}. 
In view of the developments of media (particularly the web), and advancements in mapping theories and technologies, interactive maps today are also denoted as geovisualization, which are of remarkable power to better act as a proxy for various stories.

Nevertheless, \reviseone{geovisualization for} narratives encounters several long-standing challenges,
which also appear in geovisualization\reviseone{, and even geospatial studies} in general. 
Some prominent challenges are, among others, 
the data acquisition \& integration challenge and the semantic challenge. 

The data acquisition \& integration challenge pertains to the difficulties to retrieve, process, and integrate geospatial data for narrative cartography. 
It has been commonly acknowledged that data acquisition, cleaning, integration, and apportionment cost most of the resources (such as human labors, financial resources, and time) for a typical data science project~\citep{jano2021kwgarcuser}. 
Today the data that could underlay narrative mapping are exploding, including geospatial data from authoritative sources, VGI, and other types of thematic data linked to narratives, e.g., data from Wikipedia. In this context, it is nontrivial to search, preprocess, and integrate relevant datasets in a common data format for a particular narrative mapping task. In fact, they are, among others, the 
major roadblocks of a narrative cartography project and could easily cost the majority of an entire project's resources.

An example is the development of a narrative map that describes the stories pertaining to heritage buildings in a particular city. 
The required information includes building footprints, building attribute information (e.g., building construction time, and building history description), related historical activities records, the annual tourist statistics of each building, and so on. 
However, such information often does not come from a single data repository, but instead from a number of data sources. Such dispersed data sources impose two major issues. First, data from multiple sources often require different data preprocessing strategies. Second, despite the intrinsic connections between the data from different sources, they are, in most cases, stored in the so-called "data silos". 
The reality is that leveraging these numerous data sources ``in-one-go'' is rather difficult, as the data sources are often locked away from each other with different data models, schemes, and semantics. Integrating these enormous amount of data for narrative mapping is a unique opportunity brought up in this contemporary web mapping era, and also an immense challenge.

\reviseone{Another critical challenge is the semantic challenge which} is deemed prominent in connecting maps with narratives. 
\reviseone{It can be divided into two sub-challenges - \textit{the semantic challenge of map content} and \textit{the semantic challenge in geovisualization}.}
The seminal editorial in narrative cartography by \cite{caquard2014narrative} argued that the nexuses between maps and narratives should be addressed from two perspectives, i.e., (1) maps as representations for spatiotemporal structures of stories and their relations with places, and (2) connecting maps with the mapping process through narratives. With regard to the former perspective, one core question is how to store and represent spatiotemporal data in a semantically explicit manner to facilitate narrative mapping \reviseone{-- the semantic challenge of map content.} For instance, for place names, especially the ambiguous ones such as ``Washington'', ``San Jose'' mentioned in a personal travel blog, place name recognition \citep{manning2014stanford,karimzadeh2019geotxt,wang2020neurotpr} and disambiguation \citep{overell2008using,hu2014improving,ju2016things} are indispensable preprocessing steps that link a place name to a specific geographic entity in a gazetteer \citep{goodchild2004alexandria,regalia2018gnis}. 
How to store these preprocessed data in a formalized and semantic explicit manner for narrative mapping is essential for downstream mapping tasks\reviseone{, and for 
 data reusability and map reproducibility. }

The latter perspective has been seldom addressed, and this could be partially and tacitly ascribed to the semantic challenge in geovisualization. The knowledge of geovisualization processes, i.e., how the maps are produced, and how the underlying data are transformed to graphics, is usually embedded implicitly in complex programs or in the mind of cartographers, which renders the knowledge difficult to be transferred, interpreted, expanded, and reused. Put differently, the semantics in mapping process is difficult to be represented, and formalized. 
\cite{janowicz2010semantic} regarded visualization as a sink where semantics transferred through all the components of spatial data infrastructures (SDIs) has to be aggregated, interpreted and visualized in a meaningful way. For instance, during visualization, symbols that transform underlying data to graphics bear abundant semantic information for the delivery of map content to users, and it is broadly recognized that such information should be formally represented to foster wide comprehension and reuse.\par

Over the past decades, there has been another emerging trend of the Web, namely the Web 3.0, and at its core lies the prospect of the Semantic Web \citep{berners2001semantic,bizer2011linked} that calls for a ``Web of data'' in contrast to the traditional ``Web of documents''; see the activities from the World Wide Web Consortium (W3C)\footnote{\url{https://www.w3.org/standards/semanticweb/}}. Semantic Web is partially an endeavour of formalizing and representing data on the \reviseone{web in a both human-} and machine-readable fashion to foster data interoperability, (re)usability, and applicability.
With Semantic Web technologies, a knowledge graph (KG) \citep{noy2019industry,ji2021survey} can be constructed as a data repository describing entities (e.g., places, events, and people) and their relations within or across domain(s) according to formalized ontologies, which can be seen as a directed labeled multi-graph \citep{nickel2015review}.

Thus far, a number of large-scale KGs have been constructed, including open sourced projects such as DBpedia~\citep{mendes2011dbpedia}, YAGO \citep{rebele2016yago},
and Wikidata~\citep{vrandevcic2014wikidata},  
as well as commercial projects \citep{noy2019industry}, such as Microsoft’s Satori\footnote{\url{http://www.bing.com/blogs/site_blogs/b/search/archive/2013/03/21/satorii.aspx}},
Google Knowledge Graph\footnote{
\url{https://blog.google/products/search/introducing-knowledge-graph-things-not/}
},
Facebook's social graph\footnote{\url{http://www.insidefacebook.com/2013/01/14/facebook-builds-knowledge-graph-with-info-modules-on-community-pages/}}, and eBay's Product Knowledge Graph.
These open-sourced or commercial KGs provide structured data and factual knowledge that support many intelligent applications and services such as question answering \citep{saxena2020improving,mai2019contextual}, voice assistant (e.g., Apple Siri, Amazon Alex, Google Assistant), search (e.g., Google Search, Bing Search, Amazon Product Search), and so on \citep{noy2019industry}.

Semantic Web technologies have been increasingly adopted in the geospatial domain \citep{janowicz2012geospatial,battle2011geosparql,huang2019assessment} to address some long-standing issues, such as data integration and reuse \citep[e.g.,][]{schade2012linked,huang2018synchronising}, knowledge formalization \citep[e.g.,][]{kuhn2005geospatial,huang2020towards}, and semantic interoperability \citep[e.g.,][]{janowicz2010semantic}. The increasing employment of Semantic Web technologies in the geospatial domain have fostered various geospatial KGs (denoted GeoKG hereinafter) such as LinkedGeoData \citep{stadler2012linkedgeodata}, GeoNames\footnote{\url{https://www.geonames.org/}}, and GNIS-LD \citep{regalia2018gnis}.

KGs fundamentally organize (geospatial) knowledge and data in an interlinked and formalized fashion, thereby revealing a promising avenue to (partially) resolving 
\reviseone{the data acquisition \& integration as well as the semantic challenges} 
for geospatial studies in general \citep{janowicz2012geospatial}, 
\reviseone{especially for narrative cartography. 
Specifically,} there are several advantages to use KGs for narrative cartography:
\reviseone{
\begin{enumerate}
	\item Semantic Web technologies and many existing large-scale KGs are potentially great solutions for the \textbf{data acquisition \& integration challenge}. There are many ever-growing large cross-domain KGs (e.g. Wikidata, and DBpedia), which cover a wide range of topics (e.g. events, peoples, places, and organizations) through integrating various data sources. 
	These pre-integrated cross-domain knowledge bases can potentially serve as a huge data repository for many narrative mapping tasks which can quickly reduce the data acquisition workload. 
	Moreover, they can save tremendous efforts for data integration.
Their rich context also allows map users to easily explore relations among geographic entities (e.g. places, events), and non-geographic entities (e.g., people, organizations)
	\item Modeling the underlying map data as a GeoKG is also a great practice to overcome the \textbf{semantic challenge of map content}.
	Sometimes, the existing KGs may not have the rich information that a narrative map would like to cover. So instead of only using existing KGs as the underlying map data, we can formalize our own map content as a GeoKG. This practice leads to a semantic explicit map data representation so that the KG statements about map data can be easily integrated with other KGs for various mapping or data analysis purposes.
	\item The \textbf{semantic challenge in geovisualization} can also be tackled by formalizing the geovisualization process in KGs. This is because that by encoding the geovisualization process as KG statements, we make the semantics of the mapping process explicit, which forsters better reproducibility and comprehension.
	\item The \textbf{data integration challenge} can be naturally resolved when we represent the map content in KGs, given the great power of the Semantic Web technologies in data integration, entity alignment \citep{trisedya2019entity,zhu2020collective}, and ontology alignment \citep{jain2010ontology,zhu2016spatial,zhou2020geolink}.
\end{enumerate}
}

Despite such advantages, Semantic Web technologies (and KGs) have been barely exploited for geovisualization - needless to say for narrative cartography. 
In this work, we investigate how to use \reviseone{them} 
address the aforementioned challenges in narrative cartography. 
\reviseone{Our contribution can be summarized as follow:
\begin{enumerate}
	\item To showcase how to use existing cross-domain KGs to overcome the data acquisition challenge, we developed a set of KG-based geoenrichment python toolboxes within ArcGIS Pro. 
	We show how to use these toolboxes to quickly fetch information from a popular KG (i.e., Wikidata) to make  narrative maps about historical events and trajectories directly.
	Two use cases are provided to demonstrate this idea - one for Ferdinand Magellan's expedition and the other one for the World War II.
	\item To overcome the semantic challenge of map content and the semantic challenge in geovisualization, we design a modular ontology (also denoted KG schema) for narrative cartography, which includes a map content module and a cartography module, to formally represent the map content as well as the related geovisualization process respectively.
	This modular ontology can guide the development of KGs for different narrative mapping projects.
	\item In addition, to demonstrate how we can use the cartography module to model the whole geovisualization process in a semantic explicit manner, we list some example portrayal rules which are implemented with SPARQL rules. Each of them shows how KGs connect the geovisualization process with the underlying map data explicitly.
	\item We show that our designed modular ontology is flexible enough to allow the resulting KG to link to other existing KGs, and thus substantially alleviates the efforts for data acquisition and data integration, and fosters data reusability and map reproducibility.
\end{enumerate}
}

\reviseone{The rest of this paper is structured as follows.
We first discuss some background and related works in Section \ref{sec:background}. Subsequently, in Section \ref{sec:geoenrichment}, we discuss our KG-based geoenrichment ArcGIS Pro python toolboxes and how they can help to mitigate, or even eliminate the data acquisition \& integration challenge for narrative cartography.
Next, in Section \ref{sec:odp}, we design a KG schema (i.e., ontologies) to formally model the map content as well as the related geovisualization process.
This KG schema can help to guide the development of KGs for different narrative mapping projects.
Moreover, the designed KG schema is flexible enough to allow the resulting KG to link to other existing KGs, and thus substantially alleviates the efforts for data integration, and fosters data reusability and map reproducibility.
In Section \ref{sec:conclude}, we discuss the advantages and potential limitations of our proposed KG-based narrative mapping practice with several future research directions being pointed out.
}

 \section{\reviseone{Background and Related Work}}  \label{sec:background}

\subsection{Knowledge Graphs}  \label{sec:kg_bk}
A knowledge graph (KG) provides a graph-structured way to encode facts and statements with a certain world view. From a graph view, a KG can be regarded as a directed labeled multigraph, in which a statement is composed of two entities (nodes) and a relation (a labeled, directed edge) between them. 
Accordingly, a statement in the context of the Semantic Web can be expressed in the form of a triple (\textit{h}, \textit{r}, \textit{t}), where \textit{h} is the head entity (i.e., subject), \textit{r} the relation (i.e., predicate), and \textit{t} the tail entity (i.e., object), respectively. 
For instance, a statement \textit{Santa Barbara is part of California} can be represented as (\textit{Santa Barbara}, \textit{partOf}, \textit{California}). 
Such a data model is called Resource Description Framework (RDF), a W3C standard that facilitates data integration and knowledge management among different data sources on the web.

So far, different KGs have been constructed for different purposes and different topics. For example, DBpedia, Wikidata, and YAGO are general-purpose cross-domain KGs partially built based on Wikipedia. LinkedGeoData \citep{stadler2012linkedgeodata}, GeoNames\footnote{\url{https://www.geonames.org/}}, GNIS-LD \citep{regalia2018gnis}, and GeoLink~\citep{mai2016linked,cheatham2018geolink} are GeoKGs with majorly geospatial entities. 
Bio2RDF~\citep{belleau2008bio2rdf} is a bioinformatics KG, and KnowLife~\citep{ernst2014knowlife} is a health and life science KG. 
Thanks to the RDF data model, it is straightforward to integrate data among these KGs. More specifically, all these open-sourced KGs are linked to each other, e.g., by \textit{owl:sameAs} links to indicate that two nodes from two different KGs corresponding to the same real-world entity. So these KGs jointly form a even large KG called the \textit{Linked Open Data Cloud}\footnote{\url{https://lod-cloud.net/}}, which currently has in total 1301 data repositories (KGs). These KGs, especially cross-domain KGs and geospatial KGs, provide a rich information resource for spatial analysis and geovisualization.

\subsection{Space and Time in a KG}  \label{sec:spatim_bk}
Space and time are the nexuses of knowledge representation and organization in KGs \citep{janowicz2010role}. Despite the noticeable inseparability of space and time, two terms, geographic knowledge graphs (GeoKGs) \citep{wang2019geographic,yan2019geographic,sun2020aligning} and temporal knowledge graphs (TKGs) \citep{garcia2018learning,gottschalk2018eventkg}, have become widely used in the literature. They emphasize the geographic and temporal perspectives in KGs respectively. 
In the past decade, they have contributed to a variety of geospatial studies, such as toponym resolution~\citep{grover2010use,middleton2018location}, geographic question answering~\citep{mai2019relaxing,mai2020se}, place summarization~\citep{yan2019spatially}, and travel attraction recommendation~\citep{lu2016travel}.

Table~\ref{tab:types_info_kg} summarizes the ways in which KGs represent different types of spatial and temporal information. We categorize common spatial information into four groups, including location information, spatial relations between/among them, spatial scope of statements, and non-spatial attributes about geographic entities. Meanwhile, four types of temporal information are identified, which describe the occurring time of events, temporal/non-temporal relations between events/entities, temporal scope of entities (e.g., the lifespan of a person), and temporal scope of statements, respectively. 
We also demonstrate how a KG jointly represents spatial and temporal information such as space time point or a trajectory. For each of these sub-type information, we provide triples from Wikidata/DBpedia as illustrations.
We can see that there is abundant spatial and temporal information stored in KGs. 

Those geographic entities which carry spatiotemporal information are also linked to other geographic and non-geographic entities. For example, an expedition is linked to its participants, travel origin, stopover points, and destination. The entities are also linked to other entities (i.e., through 2-degree relations). This forms an interesting graph structure that a user of narrative maps can explore.
Sometimes, to make a narrative map for some historical events or stories, one can directly get those preprocessed information from an existing KG rather than starting to collect map data from scratch. These existing KGs can be used for narrative mapping directly or can be easily integrated with additional data sources to serve as the underlining map data representation. This will significantly alleviate the data acquisition efforts. Section \ref{sec:geoenrichment} shows two examples of making narrative maps based on existing KGs.

\begin{table}[]
  \centering
  \caption{Different types of spatial and temporal information in KGs}
  \label{tab:types_info_kg}
\begin{adjustbox}{max width=1.5\textwidth, rotate=90}
    
	{\small
	
\begin{tabular}{@{}p{3.5cm}|p{3cm}|p{4cm}|p{3cm}|p{7.5cm}@{}}
\toprule
                                             & Type of Information                           & Description                                                                  & Sub-type                       & Examples                                                                                                                                                                                                                                                                                      \\ \hline
\multirow{8}{*}{Spatial Information}         & \multirow{3}{*}{Location}                     & \multirow{3}{*}{location of places, events, etc.}                            & Geometry of a place            & \textless{}New York City, geometry, POINT(-88.23 14.92)\textgreater{};                                                                                                                                                                                                                        \\ \cline{4-5} 
                                             &                                               &                                                                              & Location of an event occurring & \textless{}Pacific War, location, East Asia\textgreater{};                                                                                                                                                                                                                                    \\ \cline{4-5} 
                                             &                                               &                                                                              & Places involved in events      & \textless{}Battle of the Atlantic, participant, Karl Dönitz\textgreater{}.                                                                                                                                                                                                                    \\ \cline{2-5} 
                                             & \multirow{3}{*}{Spatial relation}             & \multirow{3}{*}{Spatial relation between places}                             & Topological relation           & \textless{}France, sharesBorderWith, Germany\textgreater{}; \textless{}Nile River, isCrossesOf, Freedom Bridge (South Sudan)\textgreater{};                                                                                                                                                   \\ \cline{4-5} 
                                             &                                               &                                                                              & Directional relation           & \textless{}Nonthaburi Province, southWest, Pathum Thani Province\textgreater{}; \textless{}River Pang, upstream, River Thame\textgreater{};                                                                                                                                                   \\ \cline{4-5} 
                                             &                                               &                                                                              & Distance relation              & \textless{}Los Angeles, isNearestCityOf, Pacific Ocean\textgreater{};                                                                                                                                                                                                                         \\ \cline{2-5} 
                                             & Spatial scope of statements                   & Declare geographic regions in which a statement is valid                     & -                              & The statement \textless{}United States of America, continent, Oceania\textgreater applies to Hawaii, Guam, American Samoa and Northern Mariana Islands only.                                                                                                                                  \\ \cline{2-5} 
                                             & Attributes about geographic entities          & Attributes used to describe a geographic entity                              & -                              & \textless{}Santa Barbara, populationTotal, 43352\textgreater{}; \textless{}Santa Barbara, areaTotal, 295600000\textgreater{};                                                                                                                                                                 \\ \hline
\multirow{5}{*}{Temporal Information}        & \multirow{2}{*}{Occurring time \reviseone{of events}}               & \multirow{2}{*}{Occurring time of events}                                    & Time instant                   & \textless{}Tunguska event, point in time, 30 June 1908\textgreater{}                                                                                                                                                                                                                          \\ \cline{4-5} 
                                             &                                               &                                                                              & Time interval                  & \textless{}World War II, start time, 1 September 1939\textgreater \& \textless{}World War II, end time, 2 September 1945\textgreater{}                                                                                                                                                        \\ \cline{2-5} 
                                             &                                               &                                                                              & Time interval                  & \textless{}World War II, start time, 1 September 1939\textgreater \& \textless{}World War II, end time, 2 September 1945\textgreater{}                                                                                                                                                        \\ \cline{2-5} 
                                             & Event relations                               & Indicate the temporal order between events or other relations between events &                                & \textless{}World War II, follows, World War I\textgreater{}; \textless{}2008 Summer Olympics, hasPart, Basketball at the 2008 Summer Olympics\textgreater{}                                                                                                                                   \\ \cline{2-5} 
                                             & Temporal scope of \reviseone{objects}                   & Indicate the life period of entities                                         & -                              & \textless{}California, Inception, 9 September 1850\textgreater{}; \textless{}Albert Einstein, dateOfBirth, 14 March 1879\textgreater{}; \textless{}Albert Einstein, dateOfDeath, 18 April 1955\textgreater{}                                                                                  \\ \cline{2-5} 
                                             & Temporal scope of statements                  & Declare the validity period of a statement                                   & -                              & The statement \textless{}United States of America, capital, New York City\textgreater is valid during the period of 1785$\sim$1790.                                                                                                                                                           \\ \hline
\multirow{2}{*}{Spatialtemporal Information} & Location and temporal information of an event & Indicate where and when an event occurs                                      & Spatialtemporal Point            & \textless{}Thomas Fire, location, Santa Barbara County\textgreater and \textless{}Thomas Fire, startTime, 4 December 2017\textgreater{}                                                                                                                                                       \\ \cline{2-5} 
                                             & Spatialtemporal information                   & Indicate the change of entities/ events in space and time                    & Trajectory                     & \textless{}Ferdinand Magellan, participantIn, Magellan–Elcano expedition\textgreater{}, \textless{}Magellan–Elcano expedition, via, Sanlúcar de Barrameda, 20 September 1519\textgreater{}, \textless{}Magellan–Elcano expedition, via, Canary Islands, 26 September 1519\textgreater{}, etc. \\ \bottomrule
\end{tabular}

} 
\end{adjustbox}

\end{table}

\subsection{\reviseone{Event}}    \label{sec:relwork-event}
\reviseone{
The objective of narrative cartography can often be linked to the concept of event in spatial information~\citep{scheider2014encoding}. To this end, previous theoretical studies on formalizing the concept of event can provide a foundation to the ontology design for narrative carography.
}

\reviseone{
As one of the earliest work on event modeling, \citet{allen1994actions} advocated the idea that \textit{events} are the way by which agents classify certain relevant patterns of changes. 
An event must involve at least one object over some stretch of time, i.e., time intervals, or involve at least one change of state. Moreover, they are defined to occur over intervals of time which cannot be reduced to some set of properties holding at instantaneous points in time. 
In contrast, \textit{actions} are something an agent (e.g., a person or robot) might do which might cause an event to occur. By associating time periods to events, \citet{allen1994actions} introduced a general representation of events and action based on the famous interval temporal logic~\citep{allen1983maintaining} which supports a wide range of reasoning tasks including planning, explanation, and prediction. 
}

\reviseone{
\citet{galton2002two} compared two event definitions from two communities -- active database and knowledge representation. 
The active database approach defines events based on their detection conditions and regards events as instantaneous. On the contrary, the knowledge representation approach defines event based on their occurrence over an interval and regards them as durative. \citet{galton2002two} showed that treating events as instantaneous is inadequate and will lead to problems during temporal reasoning. }

\reviseone{
\citet{galton2009water} compared the similarities and differences between events and objects. They pointed out that they are both discrete individuals, which cannot be dissected into parts with the same types, and have well-defined extensions. 
\citet{galton2009water} further showed that the relation between event and process can be seen as an analogy of the relation between object and matter. This neat analogy has been widely accepted for event modeling.}

\reviseone{Event is also regarded as one of the core concepts of spatial information \citep{kuhn2011core,kuhn2012core,kuhn2015designing}. Similar to previous studies, \citet{kuhn2015designing} also treated events as individual portions of processes and are temporally bounded. In many cases, events are also spatially bounded, such as wildfires, hurricanes, and floods. Additionally, they also emphasized that events have an identity as objects which are described by their temporal and thematic properties and relations.
 }

\reviseone{These previous theoretical studies on events provide a solid ground to formalize the conceptual model of events and the way how events can connect with other spatial information in a narrative cartography context. In Section \ref{sec:geoenrichment}, we show how to use existing KGs to dynamically visualize two well-known historic events -- Ferdinand Magellan's expedition and World War II as simple narrative maps with the help of a collection of KG-based GeoEnrichment toolboxes. 
Moreover, in Section \ref{sec:odp-content}, we explicitly consider the spatial temporal scoping of different geographic objects and events when we design the map content ontology.
}

\subsection{\reviseone{Semantic Web and KG Applications in Digital Humanities}}    \label{sec:relwork-humantity}

\reviseone{
The advancements in narrative cartography are also closely related to the 
``spatial turn'' in many humanities disciplines such as history, classical studies, literary studies, philology, and religion
\citep{adams2017spatial}. Except for human geography, other disciplines also began to regard space as an important dimension to their own areas of inquiry~\citep{warf2008spatial} and many humanity researchers have started to explicitly record the spatial and temporal attributes of their data and use visual analytic as part of their analysis. 
Under this ``spatial turn'' trend, narrative maps become increasingly popular in digital humanity research. One good example is Esri's Story Map collection about history, culture, literature, and the art\footnote{\url{https://collections.storymaps.esri.com/humanities/}}.
}

\reviseone{
In order to add the spatial (and temporal) dimension into the current digital humanity data repositories, a common practice is to do content annotation -- associating the spatiotemporal references, i.e., historic or contemporary toponyms, in unstructured resources to the corresponding geographic features in a gazetteer~\citep{goodchild2004alexandria,barker2016pleiades,grossner2016place}. 
A gazetteer is a geographic dictionary or directory which links place names to their geographic locations as well as other information such as place description, alternative names, feature types, their spatial relations to other places, and so on. 
Some popular gazetteers are Alexandria Digital Library Gazetteer~\citep{goodchild2004alexandria}, Pleiades Gazetteer\citep{elliott2008pleiades},  and the Geographic Names Information System (GNIS)\footnote{\url{https://www.usgs.gov/core-science-systems/ngp/board-on-geographic-names}}.
}

\reviseone{
Because of several key advantages of Knowledge Graphs and Semantic Web technologies -- improving interoperability across heterogeneous datasets, easing dataset publishing and retrieval, supporting co-reference resolution without enforcing global consistency \citep{regalia2018gnis}, we have witnessed an increasing number of gazetteers published in a Linked Data format, i.e., as a knowledge graph, to facilitate dataset discovery and integration. 
Specifically,  a number of gazetteers have been published in a Linked Data format such as Getty Thesaurus of Geographic Names (TGN)\footnote{\url{https://www.getty.edu/research/tools/vocabularies/tgn/}}, GeoNames\citep{ahlers2013assessment}, GNIS-LD~\citep{regalia2018gnis}, the Pelagios Project~\citep{barker2016pleiades}, and World Historical Gazetteer~\citep{mostern2017world,grossner2020whg}.
}

\reviseone{
One key challenge that many gazetteers encounter is how to meaningfully scope different places spatially and temporally. 
\citet{kauppinen2008creating} proposed a geospatial ontology time series to represent the meaning of changing geographic features. They proposed to represent each region with different URIs after some regional changes such as merge, split, and name change. By using this practice to encode geographic features in Finland, the part-of place hierarchy at a specific time can be automatically constructed.
\citet{grossner2016place} approached the same problem with a different approach. They proposed an ontology design pattern (ODP) for \texttt{setting}. This ODP associates each period and place with a setting which has a \texttt{SpatialScope} and a \texttt{TemporalScope}. \texttt{SpatialScope} and \texttt{TemporalScope} are the superclass of \texttt{SpatialExtent} and \texttt{TemporalExtent} accordingly. Here, \texttt{SpatialExtent} and \texttt{TemporalExtent} are models by GeoSPARL ontlology and OWL Time Ontology respectively.
}

\subsection{\reviseone{Formalizing Geospatial and Cartographic Knowledge with Ontologies}}    \label{sec:relwork-carto}

\reviseone{
Ontology is a major paradigm for knowledge representation and reasoning in Semantic Web. Specifically, ontologies are controlled vocabularies that describe concepts and relations between concepts using well-understood formal constructs; such constructs formalize the intended meaning of the vocabularies and capture background knowledge about the domain \citep{horrocks2008ontologies}. In the geospatial domain, knowledge representation using ontologies has been a long-standing research topic. 
Different ontologies have been proposed to formally represent geospatial information (e.g., vector geometries) in knowledge graphs such as NeoGeo~\citep{norton2012neogeo}, GeoSPARQL ~\citep{battle2011geosparql},  stRDF/stSPARQL~\citep{koubarakis2010modeling}, and AGO \citep{regalia2017revisiting}.
Except for modeling basic geometric information, many of the endeavours in the geospatial semantics community have been made to model more advanced geographic concepts such as trajectories \citep{hu2013geo} and sensor network \citep{janowicz2012observation}.
}

The idea of formalizing knowledge embedded in maps is intuitive in view of the implicit concepts and rules inherent in maps \citep{kavouras2007theories}. In this direction, \citet{scheider2014encoding} proposed ontologies to formally represent the content of historic maps in order to support search for map resources. \cite{varanka2018map} proposed the notion of "the map as knowledge base" (namely developing maps with GeoKGs), which is, from a technical perspective, akin to the KG-based GeoEnrichment approach for narrative mapping in this paper (cf. Section \ref{sec:geoenrichment}). Subsequently, \cite{huang2020towards,huang2020towards1, viry2021derive} took further steps to this end. They not only semantically encoded the underlying map content data, but also formalized the knowledge of visualization, i.e., how the data are transformed to graphics, so as to forge GeoKGs at both the data level (map content) and the meta-knowledge level (geovisualization theories). Besides, \cite{gao2017designing} proposed a map legend ontology to semantically annotate and query map contents via their legend in a machine-readable manner. \cite{degbelo2021ontology} formalized an ontology design pattern for map content to facilitate map interpretation and insights sharing.

These studies form a solid ground for this paper, i.e., formalizing the knowledge involved in narrative mapping. However, these previous studies predominately focused on general-purpose static visualization of geospatial data, and some of the key components in narrative mapping are lacking, e.g., in the modelling of temporal scope of events. 
In this context, we design a modular ontology tailored for narrative cartography, including the map content module and the cartography module that the narratives entail (See Section \ref{sec:odp}). Moreover, in contrast to many previous studies~\citep{scheider2014encoding,gao2017designing,degbelo2021ontology} which mainly focused on modeling map topics and content for map sharing and searching, our narrative cartography ontology formalizes the whole mapping process from the map content to the geovisualization.

\section{Knowledge Graph based GeoEnrichment for Narrative Mapping} \label{sec:geoenrichment}

\reviseone{
As we discussed in Section \ref{sec:intro}, one promising way to overcome the \textbf{data acquisition \& integration challenge} of narrative cartography is to use those pre-integrated cross-domain large knowledge graphs as the data repository for different mapping projects.
However, one question is \textit{how we can directly make use of the data from those KGs from within a Geographic Information System (GI System) to make narrative maps}. 
In fact, currently, there is no GI Systems that can directly consume Linked Data and Knowledge Graphs as one of their data formats. 
So some knowledge graph plugins need to be developed to make this possible in the first place. 
}

\reviseone{
In the following, we will show how we can use our KG-based GeoEnrichment python toolboxes to make narrative maps for different types of historical events.
Section \ref{sec:geoenrich-background} discusses some pioneer works about integrating Linked Data into GI systems. We will discuss the limitations of these KG plugins for GI Systems.
Then Section \ref{sec:geoenrich-overview} briefly discusses our KG-based GeoEnrichment Python Toolboxes for ArcGIS Pro which aim at overcoming the shortcomings of previous works.
Next, Section \ref{sec:geoenrich-exp} and \ref{sec:geoenrich-ww2} shows two use cases of our KG-based GeoEnrichment toolboxes for narrative mapping.
Section \ref{sec:geoenrichment_limit} concludes this section and discusses the limitations of our tools.
}

\subsection{\reviseone{Limitations of Existing Knowledge Graph Plugins of GI Systems}}  \label{sec:geoenrich-background}

\reviseone{The main objective of a knowledge graph plugin for a GI System is to allow a GI System to directly consume the data from many ever-growing knowledge graphs so that we can do spatial analysis or make maps on top of these KGs. }
Despite the advantages of KGs and Semantic Web technologies in modeling (spatiotemporal) data, 
few efforts have been devoted to directly consume geospatial data within KGs for spatial analysis, geoprocessing, or mapping purpose. 
\reviseone{From a 
GISystem }
perspective, Linked Data and KG research seem like a \textit{one-way street} \reviseone{\citep{mai2019deeply}}. On one hand, numerous efforts have been made to triplify the existing geospatial data into RDF triples and focus on getting various types of geo-data out of \textit{data silos}. 
On the other hand, the question of how to actually make use of this plethora of data (i.e., GeoKGs) for spatial analysis or cartography purpose remains \reviseone{largely unanswered. }
The reason is that all the current state-of-the-art GISystems and cartography softwares such as ArcGIS, QGIS, and SuperMap can not directly consume (geospatial) KGs directly without data conversion.

It is possible to flatten the graph structure of a whole KG into a table format such as Shapefiles in order to make it manipulable for GISystems. However, the converted tabular data will become another data silo and possibly get quickly out-of-date. Moreover, this data flattening practice will erase all the rich link structure provided by a KG. The resulting tabular data is very similar to those we get from the conventional web feature services (WFS) and GIS plugins such as the QGIS OpenStreetMap Plugin\footnote{\url{https://wiki.openstreetmap.org/wiki/QGIS_OSM_Plugin}}. 
\reviseone{These traditional services primarily focus on } 
fetching the geometric information as well as some basic properties (e.g., labels, types) of geographic features/entities.
Yet, from a narrative cartography perspective, we are not only interested in fetching such basic information, but also interested in \textit{exploring the relationships between places and other entities such as events, people, and activities} in which the link structure provided by KGs becomes important.

Instead of the graph flattening approach, recently we see two works - ESRI ArcMap Linked Data Connector~\citep{mai2019deeply} and QGIS SPARQL Unicorn plugin\footnote{\url{https://github.com/sparqlunicorn/sparqlunicornGoesGIS}}- that build toolboxes or plugins for existing GISystems to enable a GIS user to explore the KG structure from within a GISystem. These toolboxes or plugins can automatically construct SPARQL queries based on the user input and send them to some existing KGs, such as DBpedia and Wikidata. The results of these SPARQL queries are automatically materialized into a GIS processable format such as Shapefiles or feature classes in a File geodatabase, which can be utilized to explore the KG further or to do conventional spatial analysis. Unlike the whole graph flattening approach, these two works will not directly flatten the whole KG into a \textit{big} table. Instead, they only take a small subgraph from the KG based on users' input, convert them into a tabular format while still keep the possibility to allow users to explore the strongly interlinked graph structure further.

\reviseone{Both tools have the \textbf{area-based entity retrieval} functionality which enables}
a GIS user to define a study area and retrieve geographic entities of certain types falling into this area. They can be used to answer questions such as \textit{show me all county seats within the picked study area}. The fetched geographic entities are formatted in a Shapefile which contains information such as their Unique Resource Identifies (URIs), their feature types, their geometries, and their labels. Because of these URI information, this file can be further utilized by the same toolbox set or plugin to explore the KG further.

Despite the above neat properties, several limitations exist for these existing KG \reviseone{tools:} \begin{enumerate}
    \item \reviseone{QGIS SPARQL Unicorn plugin only support a few basic spatial query functionalities such as area-based entity retrieval. }
    They do not support more complex KG exploration and analysis (e.g., N-degree relationship exploration as shown in Figure \ref{fig:relfinder-exp-start}) which can be useful for narrative cartography.
    \item Both ESRI ArcMap Linked Data Connector and QGIS SPARQL Unicorn plugin are still geographic entity-centred. They only allow for exploring a KG starting from a geographic node. However, a user might be willing to start exploring the KG from a non-geographic entity/node, such as Ferdinand Magellan.
    \item These tools are based on out-dated GIS platforms or are poorly maintained. For instance, based on a comprehensive testing on QGIS SPARQL Unicorn plugin, 
    we find out many functionalities of QGIS SPARQL Unicorn plugin do not work. Based on the inspection on the constructed SPARQL queries, we discover some failures that are due to a systematic syntax errors in their SPARQL constructor, while others are hard to tell. As to ESRI ArcMap Linked Data Connector, although most toolboxes work well on ArcMap 10.4 - an old version of ArcGIS platform, which stopped being maintained by Esri Inc., they are not compatible with the newest ArcGIS Pro platform. \end{enumerate}

\subsection{\reviseone{Overview of our KG-based GeoEnrichment Services}}  \label{sec:geoenrich-overview}

\begin{figure}[h]
	\centering
	\includegraphics[width=1.0\textwidth]{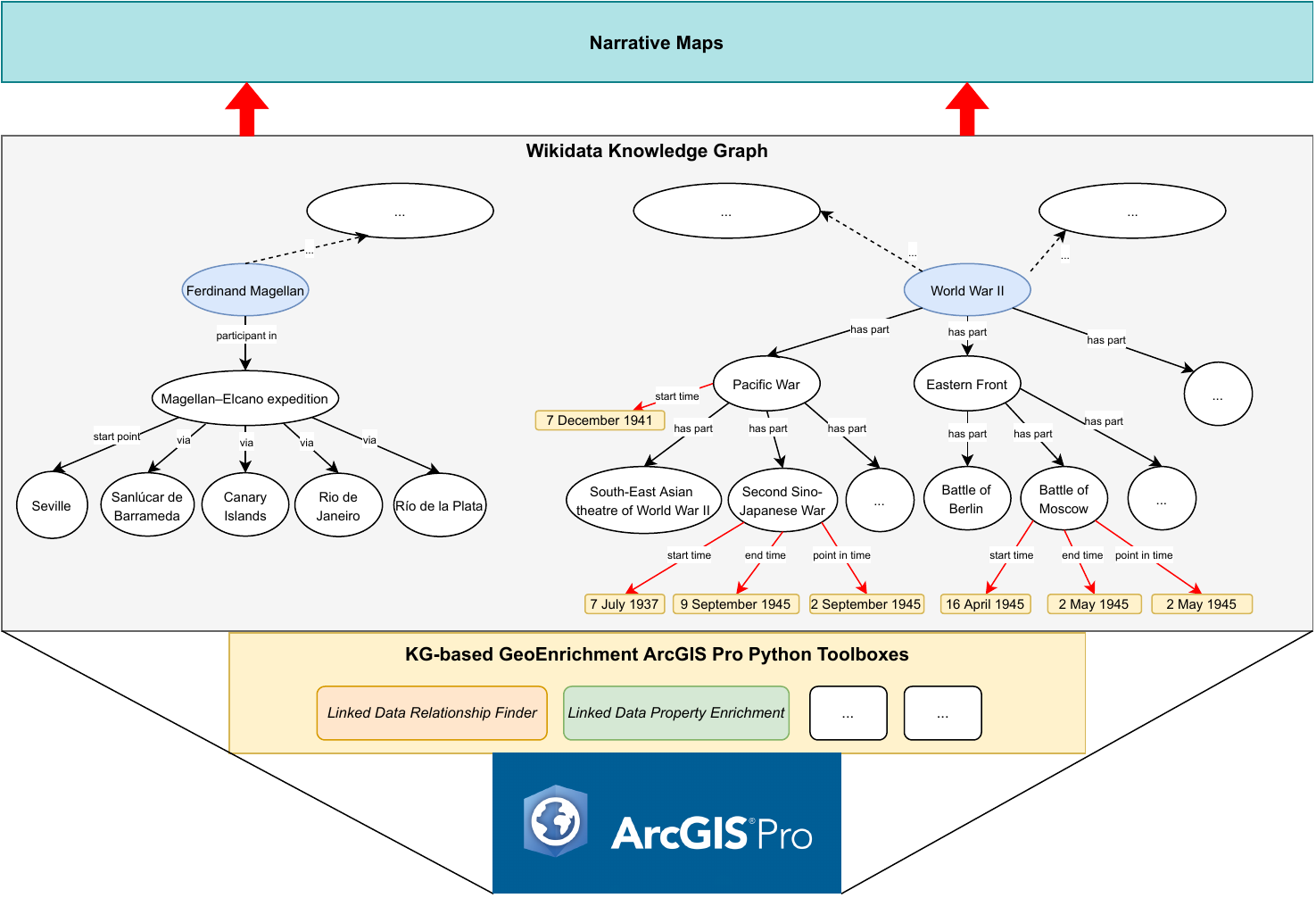}
	\caption{\reviseone{An illustration of our KG-based GeoEnrichment python toolbox workflow.}}
	\label{fig:geoenrich-workflow}
\end{figure}

\reviseone{To overcome the above limitations,}
we develop a KG oriented GeoEnrichment tool based on the ESRI ArcMap Linked Data Connector. 
It is a collection of ArcGIS python toolboxes to support narrative mapping and provide a general access to KG information from within GISystems. 
\reviseone{
Figure \ref{fig:geoenrich-workflow} illustrates how our KG-base GeoEnrichment toolset can be used for narrative mapping. 
It serves as a middle layer between a GI System, i.e., ArcGIS Pro, and a Knowledge Graph (e.g., Wikidata). A GIS user can directly explore and retrieve necessary data from the KG within ArcGIS Pro. The retrieved data is materialized in a GIS format (i.e., Shapefiles and ArcGIS Attribute Tables) so that normal spatial analysis and cartography operations can be applied on them.
We can directly make narrative maps based on these data while the time and efforts for data retrieval, preprocessing, and integration are largely reduced.
Compared with those two previous tools, our new KG-based GeoEnrichment toolset have the following advantages: 
1) It supports more flexible KG exploration and data retrieval functionalities such as N-degree relation exploration and non-geographic entity property enrichment;
2) It allows users to explore KGs from a non-geographic node; 
3) It is developed for the newest ArcGIS Pro which is rather easy to maintain.
}

\reviseone{Compared with the ESRI ArcMap Linked Data Connector, our new KG-based GeoEnrichment toolset is mainly different in two toolboxes:
\begin{enumerate}
	\item \textit{Linked Data Relationship Finder}: This toolbox enables users to explore N-degree relationship paths within a KG such as Wikidata from within a GI System.
The idea of N-degree relation exploration is shown in Figure \ref{fig:geoenrich-workflow}. 
	This toolbox requires several input parameters: the SPARQL endpoint, start node(s), relationship degree, and the property direction and property URL of the $K$th degree property along the property path.
	The SPARQL endpoint is the SPARQL endpoint of the KG a user would like to connect to. Currently, our tool supports Wikidata as well as any other KGs who support GeoSPARQL.
	The start nodes are the user selected nodes where our property path begins which are denoted as blue nodes in Figure \ref{fig:geoenrich-workflow}. 
	ESRI ArcGIS Linked Data Connector~\citep{mai2019deeply} only allows geographic entities as the start nodes while we relax this restriction and allow non-geographic node as start nodes, e.g., \texttt{Ferdinand Magellan} or \texttt{World War II}. 
	In addition, a user need to let the tool know which property path (s)he want to explore. This includes
	the relationship degree (the length of the property path), each property's URL as well as its direction along the path.
	As shown in Figure \ref{fig:geoenrich-workflow}, when a user specifies \texttt{Ferdinand Magellan} as the start node, \texttt{?people} $\rightarrow$ \texttt{participant in} $\rightarrow$ \texttt{?expedition} $\rightarrow$ \texttt{via} $\rightarrow$ \texttt{?place} as a 2-degree property path\footnote{Here, we use ``?'' to indicate a variable or placeholder for the entities/nodes along a property path. Property \texttt{participant in} and \texttt{via} are the first degree and second degree relations.}, all the expeditions taken by Ferdinand Magellan and all the places these expeditions have past will be retrieved.
	Figure \ref{fig:relfinder-exp-start}, \ref{fig:relfinder-exp-via} and \ref{fig:relfinder-ww2} show how this toolbox looks like in different use cases. 
	Note that this toolbox adopts an interactive way to allow users to specify the property path as \citet{mai2019deeply} did. For example, when a user select \texttt{Ferdinand Magellan} as the start node and \textit{ORIGIN} as the first relation direction, a SPARQL query will be constructed to get all properties that have \texttt{Ferdinand Magellan} as its subject node. The rest works in a similar manner. 
	\item \textit{Linked Data Property Enrichment}: This toolbox allows a user to \textit{enrich} the retrieved data with more information from the KG. For example, when we retrieve all the events which are transitatively part of \texttt{World War II} based on Linked Data Relationship Finder toolbox as shown in Figure \ref{fig:geoenrich-workflow}, we can enrich these events with more attributes such as \texttt{start time}, \texttt{end time}, and \texttt{point in time} which are indicated as red arrows. 
	ESRI ArcGIS Linked Data Connector~\citep{mai2019deeply} also provided a similar property enrichment toolbox. However, their toolbox only allows geographic entities as the input entities for property enrichment. This will post a lot of limitations on the kind of information we can retrieve from a KG. In contrast, our toolbox allow property enrichment for non-geographic entities. Figure \ref{fig:propenrich-ww2} is a screenshot of this tool.
\end{enumerate}
}

\reviseone{
In the following, we use two use cases to demonstrate how we can use the KG-based GeoEnrichment toolboxes in narrative mapping.
}

\subsection{A Map of Ferdinand Magellan's expedition}   \label{sec:geoenrich-exp}

A typical narrative map example is to map people's activities across space and over certain time period such as tracing the movements of characters within James Joyce’s Dubliners \citep{joyce2008dubliners}. Here, we showcase a narrative map concerning the Ferdinand Magellan's expedition in the 16th century. 

To make such a map, in addition to the trajectory of this expedition, we are also interested in exploring the people - expedition - place relationships which can not be done through traditional web feature services. A typical narrative mapping practice is to read the historical record of Ferdinand Magellan such as his Wikipedia page and then map out the expedition's trajectory. This requires to recognize the place names appeared in this article, link them to the corresponding geographic entities in an existing gazetteer, find out the geographic locations of these places as well as when Ferdinand visited them, and finally prepare a Shapefile based on such information for narrative mapping. However, with our KG-based GeoEnrichment tool, we can make such map within a few minutes from within ArcGIS directly.

\begin{figure}[h]
\centering \tiny
    \begin{subfigure}[b]{0.487\textwidth}
        \centering
        \includegraphics[width=\textwidth]{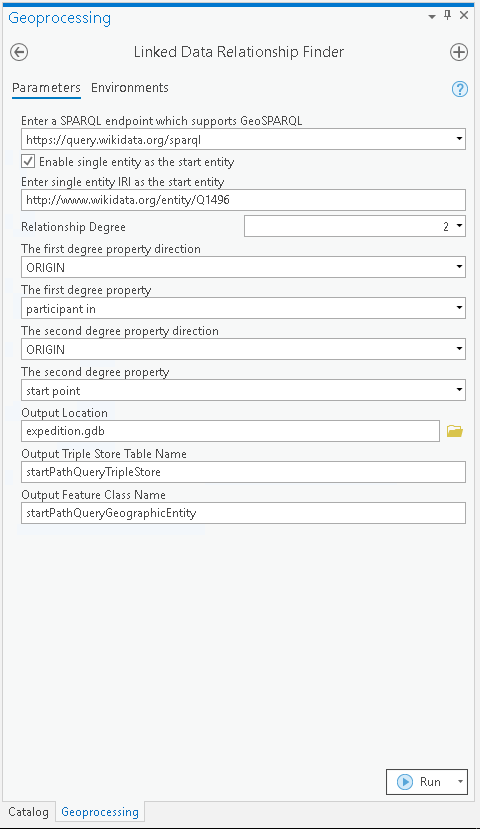}
        \caption[]{{Relationship Finder: find all expeditions and their start points
        }} 
        \label{fig:relfinder-exp-start}
    \end{subfigure}
    \hfill
    \begin{subfigure}[b]{0.49\textwidth}
        \centering
        \includegraphics[width=\textwidth]{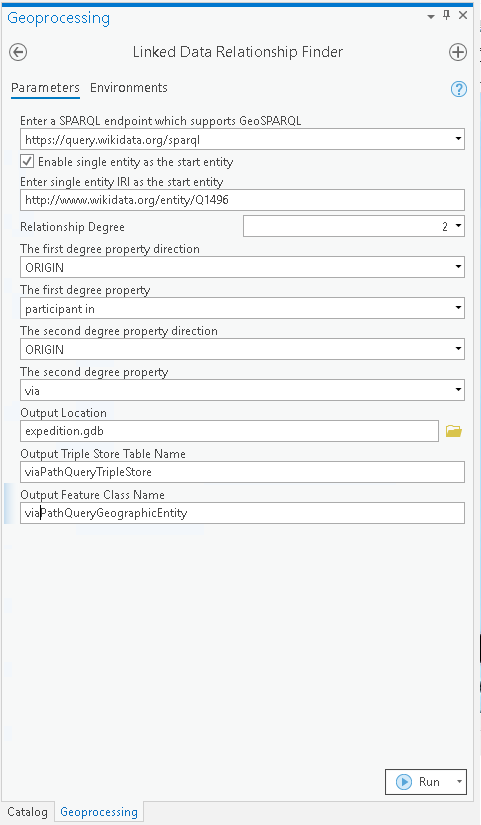}
        \caption[]{{Relationship Finder: find all expeditions and their stopover points (via point)
        }} 
        \label{fig:relfinder-exp-via}
    \end{subfigure}
    \hfill
    \begin{subfigure}[b]{1.0\textwidth}
        \centering
        \includegraphics[width=\textwidth]{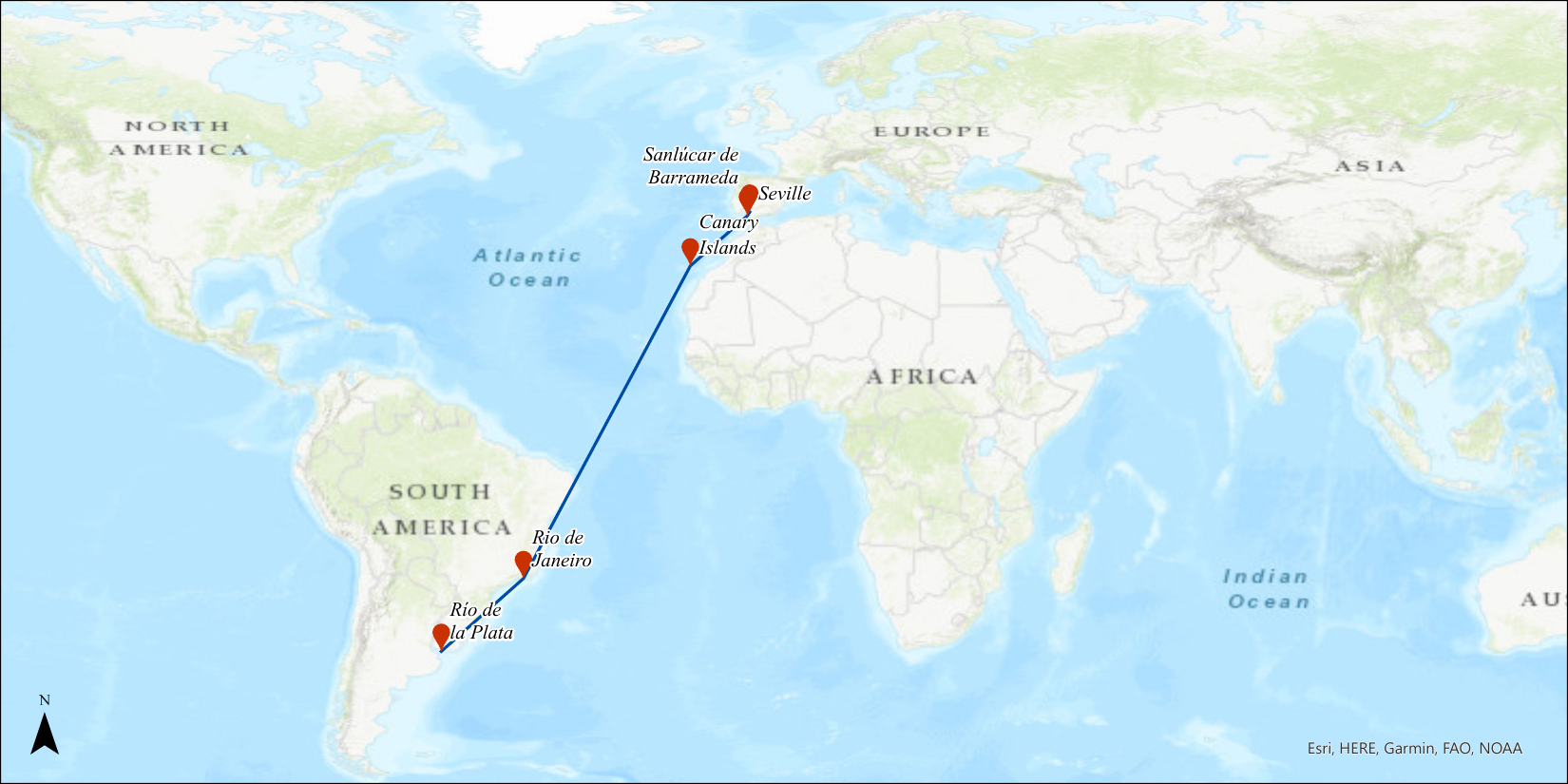}
        \caption[]{{Ferdinand Magellan's expedition
        }} 
        \label{fig:exp-map}
    \end{subfigure}
\caption{
A use case of KG-based GeoEnrichment tool: mapping the Ferdinand Magellan's expedition in the 16th-century.
We use the \textit{Relationship Finder} toolbox to find all expeditions participated by Ferdinand Magellan and all these expeditions' (a) start points and (b) stopover points (via).
(c) The resulted map shows the Ferdinand Magellan's expedition in the 16th-century.
Here, \url{http://www.wikidata.org/entity/Q1496} indicates Ferdinand Magellan in Wikidata.
}
\label{fig:geoenrich-exp}
\end{figure}

Figure \ref{fig:geoenrich-exp} shows the whole process as well as the resultant map. 
\reviseone{Firstly, Figure \ref{fig:relfinder-exp-start} shows how we can use the \textit{Linked Data Relationship Finder} toolbox to explore the property path -- \texttt{?people} $\rightarrow$ \texttt{participant in} $\rightarrow$ \texttt{?expedition} $\rightarrow$ \texttt{start point} $\rightarrow$ \texttt{?place}.}
In this case, we start from Ferdinand Magellan\footnote{\url{http://www.wikidata.org/entity/Q1496}} (the entity \texttt{wd:Q1496} in Wikidata) and explore its 2-degree relationship paths. 
The particular example path shown here goes from the person (Ferdinand Magellan) to the expeditions he took (e.g., \texttt{Magellan–Elcano expedition}\footnote{\url{http://www.wikidata.org/entity/Q1225170}}) through the \texttt{participant in} relation, and then to the start points of these expeditions (e.g., \texttt{Seville}\footnote{\url{http://www.wikidata.org/entity/Q8717}}) through the \texttt{start point} relation. 
Similarly, Figure \ref{fig:relfinder-exp-via} 
explores the people-expedition-via point (stopover points) relationship paths. The resulting start points and stopover points are automatically materialized into a Shapefile format so that we can directly map them as a trajectory.

Figure \ref{fig:exp-map} shows the resultant map of Ferdinand Magellan's expedition. Note that this map is based on the available information on Wikidata and may not reflect the whole trajectory of this expedition. 
\reviseone{This section focuses on the question how to utilize KGs to overcome the data acquisition \& integration bottleneck for narrative cartography, }
rather than producing a real ready-to-go narrative map product. 
This map shown in Figure \ref{fig:exp-map} is simply a use case and should not be judged from an aesthetic aspect. 
Nevertheless, we believe that the promotion of open-sourced KGs and cartography with KGs are mutually beneficial. Since the more people can find ways to utilize data in open-sourced KGs (e.g., Wikidata in this example), the more incentive they have to contribute to these KGs.

\subsection{A Map of All Events During World War II}   \label{sec:geoenrich-ww2}

Another example is mapping the major events happening in World War II (WW2) which might be interesting for a student or a historian. 
Since the events in Wikidata are organized in a hierarchical way, we explore the event - subevent - subsubevent relationships to obtain all events' locations as well as their temporal sequence during WW2. Here, we show how to use our KG-based GeoEnrichment tool to easily do this.

\begin{figure}[ht!]
\centering \tiny
    \begin{subfigure}[b]{0.49\textwidth}
        \centering
        \includegraphics[width=\textwidth]{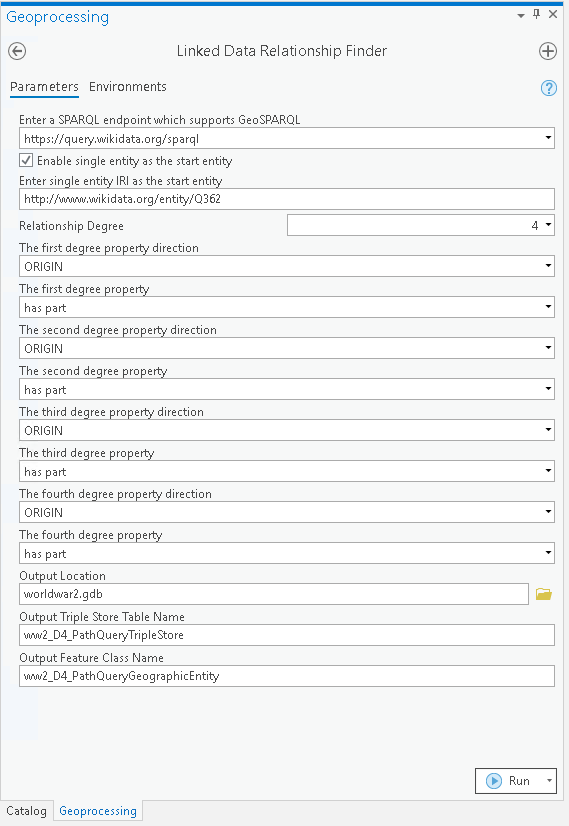}
        \caption[]{{Relationship Finder: find all events during World War II
        }} 
        \label{fig:relfinder-ww2}
    \end{subfigure}
    \hfill
    \begin{subfigure}[b]{0.49\textwidth}
        \centering
        \includegraphics[width=\textwidth]{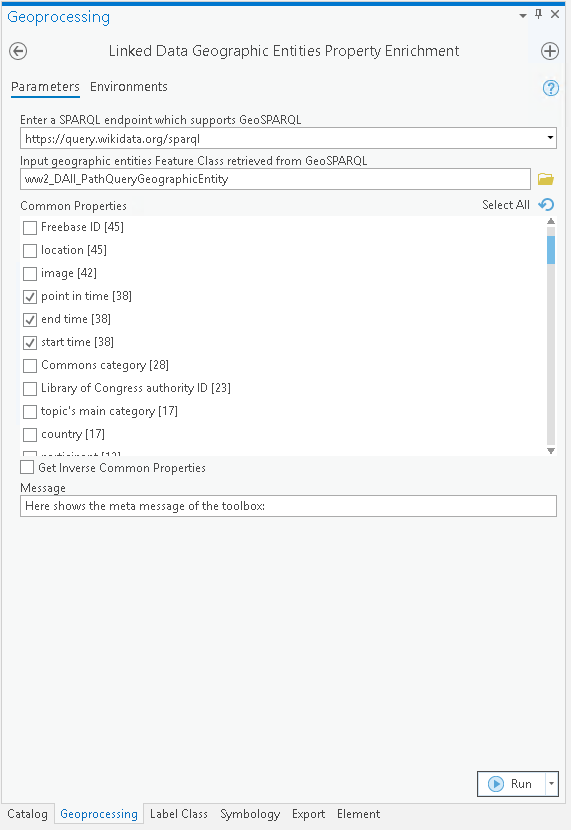}
        \caption[]{{Property Enrichment: find the time information of all events during World War II
        }} 
        \label{fig:propenrich-ww2}
    \end{subfigure}
    \hfill
    \begin{subfigure}[b]{1.0\textwidth}
        \centering
        \includegraphics[width=\textwidth]{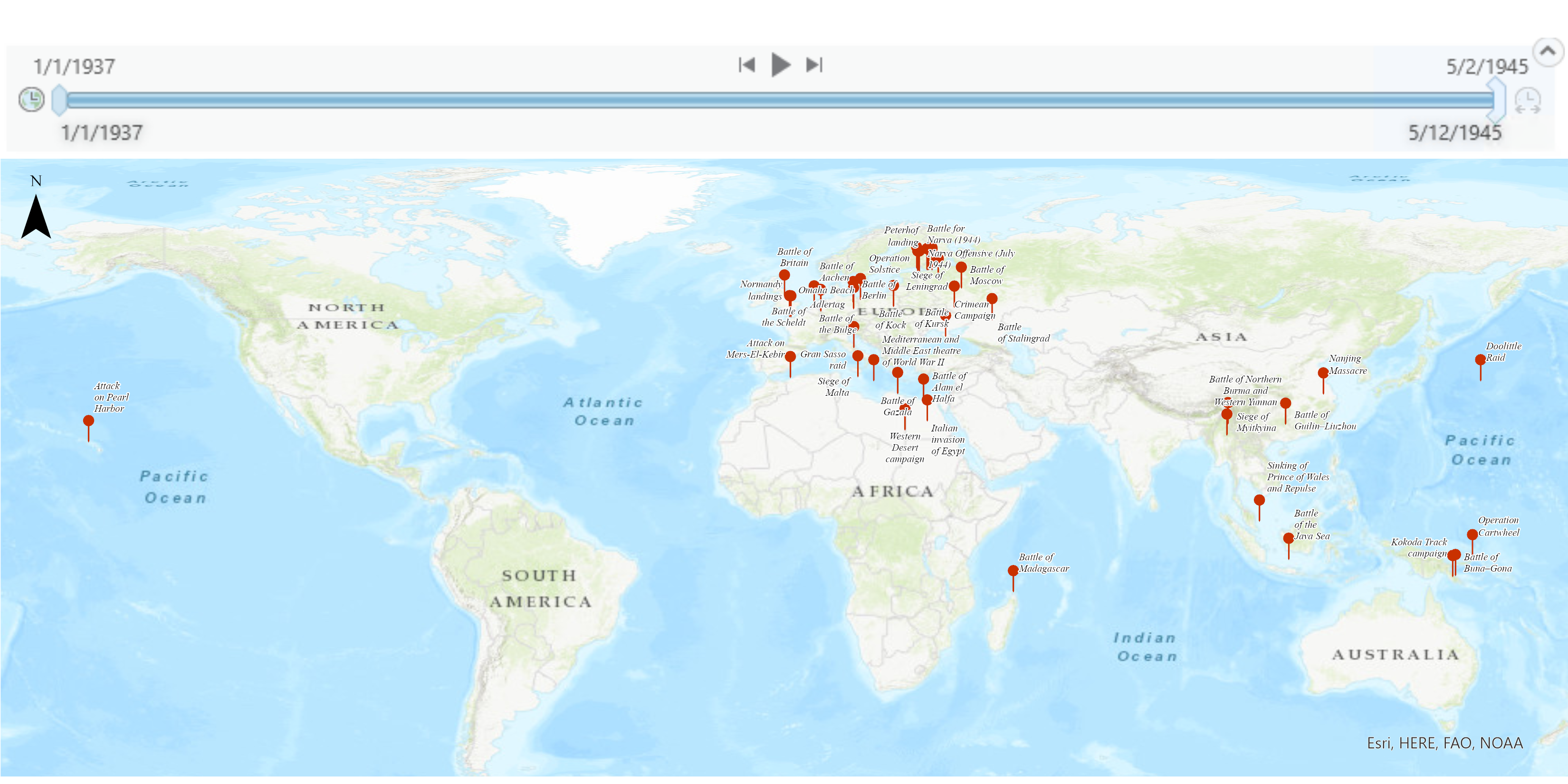}
        \caption[]{{World War II Map
        }} 
        \label{fig:ww2-map}
    \end{subfigure}
\caption{
A use case of KG-based GeoEnrichment tool: mapping all events happened during World War II.
(a) We use the \reviseone{\textit{Linked Data Relationship Finder}} toolbox to find all events that is transitively part of \texttt{World War II} (maximum 4 degree away) with \texttt{has part} relation and get all their names and geographic locations. The result is materialized as a Shapefile.
(b) We use the \reviseone{\textit{Linked Data Property Enrichment}} toolbox to query for more information about these events such as their \texttt{start time}, \texttt{end time}, as well as \texttt{point in time} which serve as their temporal information.
(c) The resulted map shows all events during World War II which are ordered by the timeline shown on the top of the map.
Here, \url{http://www.wikidata.org/entity/Q362} indicates \texttt{World War II} in Wikidata.
}
\label{fig:geoenrich-ww2}
\end{figure}

Figure \ref{fig:geoenrich-ww2} shows how to use the \textit{Linked Data Relationship Finder} tool to explore 4-degree relationship paths involving
 \texttt{?event} $\rightarrow$ \texttt{has part} $\rightarrow$ \texttt{?subevent}
relations\footnote{Words with upper case indicates a class or entity type} with \texttt{World War II}\footnote{\url{http://www.wikidata.org/entity/Q362}} as the start node. 
Here, each \reviseone{\texttt{EVENT}} node is connected to its direct \reviseone{\texttt{SUBEVENT}} node with \texttt{has part} relation.
Figure \ref{fig:geoenrich-workflow} shows the subgraph of Wikidata in which \texttt{World War II} is connected to the first and second degree of its subevents.
All these events which are transitively part of \texttt{World War II} \reviseone{(maximum 4 degree away)} are retrieved from Wikidata and materialized as a Shapefile for geovisualization.

Since we are also interested in temporal order of these events, we can enrich this GIS data by querying the temporal information of these events from Wikidata. Figure \ref{fig:propenrich-ww2} shows how we can do this based on another toolbox - \textit{Linked Data Property Enrichment} toolbox. After loading the materialized GIS data into the toolbox, it will automatically query for the properties of these events that are available for data enrichment. Here, we pick \texttt{start time}, \texttt{end time}, \texttt{point in time} as they indicate the temporal information of these events. Figure \ref{fig:ww2-map} shows the final map of these events during World War II. The timeline above can also control which events to show based on the retrieved timestamps.

Another important advantage of using KG for narrative cartography is that KGs contain massive information about each geographic and non-geographic entity. When we visualize some geographic entities (e.g., events, objects) on the narrative map, KGs can provide rich contextual information for these entities and allow the users to do further exploration.
Since many existing KGs such as Wikidata are built based on various data sources, it contains massive amount of information for each entity which will be much more than what we normally get from a single dataset (e.g., a historic battle dataset). For instance, by using the \textit{Linked Data Property Enrichment} toolbox, we can not only access the temporal information but also numerous other types information about this event such as the event category, the number of deaths, the number of injured, significant events during these events, participants, the cause of this event, following events, and so on. 
More specifically, as shown in Figure \ref{fig:propenrich-ww2}, there are 76 different properties shown in the multi selection box that are available to enrich the retrieved WW2 event dataset.  The exact SPARQL query result (i.e., all the available event properties) can be accessed with this link\footnote{\url{https://api.triplydb.com/s/0-0rfTBC_}}. 
These additional properties provide rich contextual information for each event during World War II.
In contrast, a traditional interactive map for World War II such as \textit{World War II Interactive Map}
\footnote{\url{https://ww2db.com/map/interactive/}} only contains a short description for each event.

Note that all those steps discussed above can be accomplished in a few minutes through interactions with the GIS platform.
In contrast, if we want to follow the traditional narrative cartography practice, it will require a lot of efforts make such map 
since those event information sources are scattered in different parts of a narrative and substantial efforts are needed to extract these information, preprocess them, then make such information ready for a cartography program. 
However, the event information is readily available through a KG such as Wikidata. This will substantially reduce the data acquisition efforts and accelerate the mapping process. Moreover, if one would like to add more information which is missing from the Wikidata, it is also very easy to integrate these repositories (KGs) together given the power of the underlining Semantic Web technologies.

\subsection{\reviseone{Limitations of the KG-based GeoEnrichment Approach for Narrative Cartography}}   \label{sec:geoenrichment_limit}

Despite these advantages of the KG-based GeoEnrichment \reviseone{toolboxes}, we also observe several limitations \reviseone{if we only use existing KGs for narrative mapping with the help of our toolboxes. For specifically, there are three challenges -- data incompleness of the existing KGs, semantic challenges in modeling map content, and semantic challenges in modeling the geovisualization process.}

The first limitation is the data incompleteness of the existing KGs for the select map topic.
For example, for World War II use case in Section \ref{sec:geoenrich-ww2}, we \reviseone{retrieved} 48 unique significant events who are transitively part of \texttt{World War II} in Wikidata. However, \textit{World War II Interactive Map}
\footnote{\url{https://ww2db.com/map/interactive/}} shows 334 different events. One of the reasons for this discrepancy is that 
in Figure \ref{fig:geoenrich-ww2} we only consider 1-, 2-, 3-, and 4-degree \texttt{has part} relationship/property paths from the \texttt{World War II} node (\textit{wd:Q362}) while subevents that are more than 4 degree away from \texttt{World War II} node with \texttt{has part} relation will not be retrieved. For example, \texttt{Battle of Mount Song}\footnote{\url{http://www.wikidata.org/entity/Q13403439}} (wdt:Q13403439) is a famous battle during War World II which is 5-degree away from the \texttt{WW2} node. 
However, even if we account for all the sub-events which is x-degree way from the WW2 node,  
by using this property path query\footnote{\url{https://api.triplydb.com/s/SFeGPosa4}}, we can only retrieve 122 subevents.

Another important reason is the data incompleteness issue of Wikidata. 
Wikidata, as one of the world largest collaborative open-sourced KGs, still suffers from the data incompleteness issue. It is possible that some events during the World War II are missing (entity missing) and some links among these events are missing from Wikidata (link missing). 
As for entity missing, 
one example in the World War II use case is that one specific assault of the Japanese Army during the Defense of Sihang Warehouse\footnote{\url{https://en.wikipedia.org/wiki/Defense_of_Sihang_Warehouse}}  is not instantiated as an event node in Wikidata. 
Compared with entity missing, link incompleteness is more common in existing KGs. 
For example, the \texttt{Second Sino-Japanese War}\footnote{\url{http://www.wikidata.org/entity/Q170314}} is not linked to the \texttt{First Battle of Changshan}\footnote{\url{http://www.wikidata.org/entity/Q709333}} with a \texttt{has part} relation while they are linked reversely with a \texttt{part of} relation despite the fact that these two relations are inverse to each other. This indicates that there are missing links among these event entities.
So in order to find all subevents of World War II, we should use the SPARQL query shown in Listing \ref{q:ww2_subevent} which combines two property path query patterns (1) and (2) with the \texttt{has part} and \texttt{part of} relation. This query\footnote{\url{https://api.triplydb.com/s/EaPJYM1O8}} will return 2087 events as the result which is far more than those listed in \textit{World War II Interactive Map}. This also shows the limitation of our KG-based GeoEnrichment tool to support various property path queries. We leave this as a future work and discuss it in Section \ref{sec:conclude}.

\vspace{0.5cm}
\hspace{-0.5cm}
\begin{minipage}[c]{\columnwidth}
	\begin{lstlisting}[
	linewidth=\columnwidth,
	breaklines=true,
	basicstyle=\ttfamily, 
	captionpos=b, 
	caption={Query for all sub-events of World War II in Wikidata. 
	}, 
	label={q:ww2_subevent},
	frame=single
	]
@[prefix definitions]
SELECT distinct ?event ?eventLabel 
WHERE
{
  {wd:Q362 wdt:P527* ?event .}   # (1) Find all events who is x-degree away from the WW2 node with ``has part'' relation 
  union
  {?event wdt:P361* wd:Q362 .}   # (2) Find all events who is x-degree away from the WW2 node with reverse ``part of'' relation 
  
  # Get the label for each event
  SERVICE wikibase:label {       
     bd:serviceParam wikibase:language "en" .
   }       
}
	\end{lstlisting}
\end{minipage}

Another limitation is semantic incompatibility. It can happen within the existing KG or between the existing KGs and the intended map topic.
The former means there are some semantic conflicts within a single KG. We take the World War II use case again to demonstrate this. Based on the description in Wikidata, World War II started on September 1st, 1939 (\texttt{start time}) and ended on September 2, 1945 (\texttt{end time}). So strictly speaking, all events outside of this time interval should not be consider as its subevents. However, as shown in Figure \ref{fig:ww2-map}, the timeline starts on January 1st, 1937 because Wikidata also asserts the \texttt{Second Sino-Japanese War} which happened on January 7th, 1937 as a subevent of \texttt{World War II}. This is a common data disagreement issue in open-source KGs because of the open world assumption.
The latter means that the semantics of a concept such as \texttt{World War II} in the existing KGs might be different from the one that cartographers have in their mind.
For example, \texttt{Jinan Incident}\footnote{\url{http://www.wikidata.org/entity/Q709855}} happened on May 11th, 1928 is considered as an event of WW2 in the \textit{World War II Interactive Map} (See this link\footnote{\url{https://ww2db.com/battle.php?q=China&list=c}}). However, Wikidata does not declare such a subevent relation (i.e. \texttt{part of} relation). According to the open world assumption \citep{drummond2006open} adopted by the Semantic Web technologies, we are not able to make any assertion relationship between this event and World War II. However, given the fact that \texttt{Jinan Incident} happened 11 years before the defined start time of World War II, instead of assuming that the link between them is missing because of data incompleteness, we are inclined to believe that \texttt{Jinan Incident} is treated as an individual event, rather than a subevent of the \texttt{WW2} in Wikidata. This indicates a semantic incompatibility between Wikidata and \textit{World War II Interactive Map}. To solve the semantic incompatibility issue within one KG, we can define some data quality constraints (e.g., with the SHACL \reviseone{shapes}\footnote{\url{https://www.w3.org/TR/shacl/}}). With regard to the semantic incompatibility issue among different existing KGs and \reviseone{the intended map topic, }
we need to formally define an ontology for map content to enable the semantic interoperability among them. We will demonstrate an ontology design pattern for map content in Section \ref{sec:odp-content}.

The third and last limitation comes from the semantic challenges in geovisualization discussed in Section \ref{sec:intro}. Although our KG-based GeoEnrichment tool can allow a GIS user to explore existing KGs within a GISystem, it only concerns about the map content but not the geovisualization process. How to interpret the data retrieved from an existing KG and how to visualize them on the map would depend on cartographers. The question on whether to visualize a battle as a circle, a drop pin, or a cross on the map is implicitly embedded into the mind of cartographers. To explicitly express the semantics of the geovisualization process, we need to design an ontology design pattern for it which will be discussed in Section \ref{sec:odp-carto}.

 \section{A Modular Ontology for Narrative Cartography} \label{sec:odp}

To tackle the \reviseone{semantic challenges in modeling map contents as well as geovisualization process} discussed in Section \ref{sec:geoenrichment_limit}, 
in this section, we take a step forward in the direction of underpinning narrative cartography with KGs, i.e., we formalize the knowledge involved in producing narrative maps in several ontologies. Such a formalization work is divided into two parts, namely the narrative map content ontology, and the data visualization (cartography) ontology. \reviseone{These ontologies are available through our Github repository\footnote{\url{https://github.com/RightBank/narrative-cartography-ontology}}}

\subsection{The Map Content Module}   \label{sec:odp-content}

In principle, the types of map content used in narrative cartography are numerous; apart from its main body (the map and its legend), there can also be other types of media such as image, audio, and video to serve as attachments of the maps. This implies that the representation forms in narrative cartography are both various and uncertain - it is uncertain which forms are employed in a particular context. Therefore, in this paper we concentrate on the formalization of the main body of the map content, i.e. the map itself but not its attachments. 

\begin{figure}[h]
	\centering
	\includegraphics[width=1.0\textwidth]{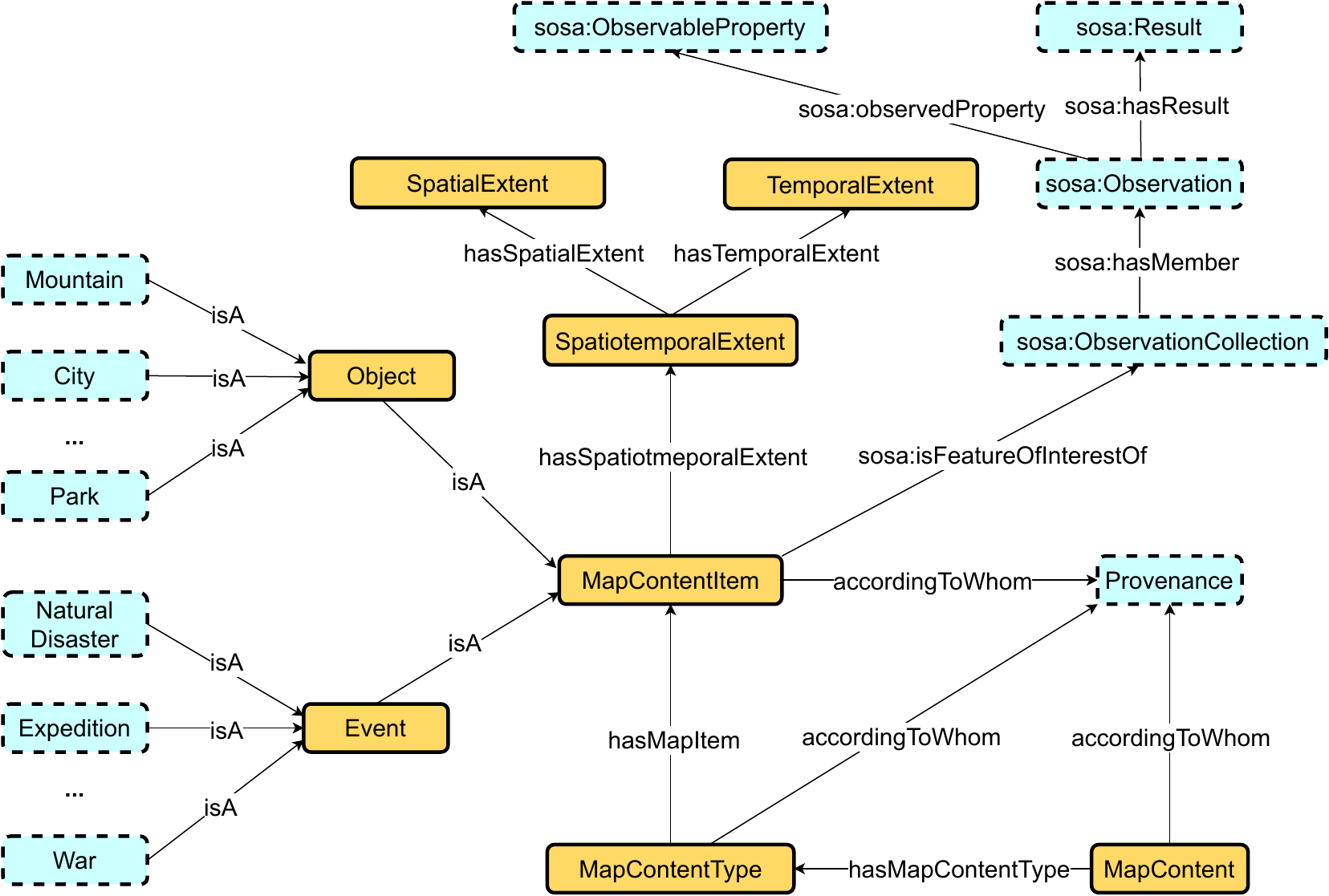}
	\caption{The \textit{Map Content} module of the \textit{Narrative Cartography Ontology}.}
	\label{fig:content-odp}
\end{figure}

\reviseone{The main objective for designing \textit{Map Content} module is to overcome the semantic challenge in modeling map content and fix the semantic incompatibility among different data sources. }
The designed \textit{Map Content} module in the narrative cartography ontology is demonstrated in Figure \ref{fig:content-odp} (note that the concepts shaded in orange are in this module, and others are reused from other ontologies). In this ontology, the most generalized concept is \texttt{MapContent}, representing the entire content of narrative maps to be rendered. An instance of \texttt{MapContent} is associated with one or several \texttt{MapContentType} through the relation \texttt{hasMapContentType}. The concept \texttt{MapContentType} generalizes the phenomenon that the narrative entails and is to be exhibited with cartography. An instance of \texttt{MapContentType} can be interpreted as a map layer used in a narrative map, e.g., the collection of events occurred during World War II can be viewed as an instance of this concept. 
One \texttt{MapContentType} consists of several \texttt{MapContentItem} with the \texttt{hasMapItem} relation. An instance of \texttt{MapContentItem} is one individual item (e.g., a particular battle during World War II) needs to be display on the map.

With regard to narrative cartography, there are two major types of geographic entities need to be visualized - \texttt{Object} and \texttt{Event}. They can be further classified into different subclass. \texttt{Object} can be classified into \texttt{Mountain}, \texttt{City}, \texttt{Park}, and so on while \texttt{Event} can be classified into \texttt{Natural Disaster}, \texttt{Expedition}, \texttt{War}, and etc. These object and event classifications depend on the real map content and several existing geographic feature type classification schema can be used here such as the Geographic Names Information System (GNIS) Feature Classification schema \citep{regalia2018gnis}. In other words, these object and event classifications can be borrowed from other ontology design patterns and we indicate them in blue boxes with dash line boundaries.

Each instance of \texttt{MapContentItem} should be associated with an instance of \texttt{Spatiotemporal- Extent}, which is used to describe the spatiotemporal scope of an object or an event. 
\texttt{Spatio- temporalExtent} is further divided into \texttt{SpatialExtent} and \texttt{TemporalExtent} to separately delineate the spatial scope and temporal scope of the associated instance of \texttt{MapContentItem}.
For instance, for an event - \texttt{Battle of Sedjenane}, its temporal scope is during February 26 to March 4 of 1943, and its spatial scope is a Tunisian town ``Sejenane'' (according to the relevant entity in Wikidata\footnote{\url{https://www.wikidata.org/wiki/Q4872340}}). 
An object can also have a temporal scope. For example, \texttt{Soviet Union} has a temporal scope from December 30th, 1922 to December 26th, 1991.
Note that the use of \texttt{SpatialExtent} is to visualize the spatial footprints of events and objects. It corresponds to the location information (the 2nd row) discussed in Table \ref{tab:types_info_kg}. 
The use of \texttt{TemporalExtent} facilitate the formalization of temporal sequence of different \texttt{MapContentItem} for the geovisualization of narratives. It corresponds to the occurring time and temporal scope of entities in Table \ref{tab:types_info_kg} under the class of temporal information. 
\texttt{TemporalExtent} is particularly important for narrative cartography compared to other geovisualization tasks since the temporal ordering of the events and stories are the focus of it.

It is also valuable to model the ``observations'' of each object and event. Here, ``observations'' means the observed properties that are used to describe objects and events. We use \texttt{sosa:Observation} from the SOSA ontology \reviseone{\citep{janowicz2019sosa}} as well as \texttt{sosa:Observation- Collection} from the SOSA extension~\footnote{\url{https://www.w3.org/TR/vocab-ssn-ext/}} to model them.
The attributes about geographic entities in Table \ref{tab:types_info_kg} can be modeled as ``observations''.
For example, the elevation of \texttt{Rio de Janeiro}, one stopover of Ferdinand Magellan's expedition, can be modeled as an instance of \texttt{sosa:Observation} with \texttt{elevation above sea level} as its \texttt{sosa:ObservableProperty} and the elevation number as its observation result - \texttt{sosa:Result}. If there are multiple observations for the same event or object (e.g., population, elevation, precipitation, and etc), we can model them as different \texttt{Observation} instances which are all linked to one \texttt{ObservationCollection} instance. This \texttt{ObservationCollection} instance is then link to the corresponding \texttt{MapContentItem} instance through \texttt{sosa:isFeatureOfInterestOf} relation.

In order to enrich the metadata of the target instances of \texttt{MapContentItem}, \texttt{MapContentType}, and \texttt{MapContent}, they can be associated with provenance information, e.g., by using the PROV Ontology\footnote{\url{https://www.w3.org/TR/prov-o/}}).

Note that the \textit{Map Content} module shown in Figure \ref{fig:content-odp} is a general ontology design pattern (ODP) to model the content of a narrative map. A cartographer can build a map content KG based on this ODP. Since this ODP is very general and flexible, it can be easily linked to some existing KGs such as Wikidata and DBpedia (i.e., data integration) based on entity alignment between \texttt{MapContentItem} instances in this KG and geographic entities in other existing KGs.
After entity alignment step, the map content KG will be significantly enriched with other data repositories. This will lead to a more powerful geovisualization which allows users to explore all these rich contextual information. Based on this formally defined ODP and entity alignment, we can also solve the \textit{semantic incompatibility} issue (see Section \ref{sec:geoenrichment_limit}) and achieve semantic interoperability among different KGs.

\subsection{The Cartography Module}   \label{sec:odp-carto}

The portrayal of the instances of \texttt{MapContentType} is also formalized into an ontology module, namely the \textit{Cartography} module (Figure \ref{fig:content-odp}; the concepts shaded in green are in this module). At the most generalized level, the concept \texttt{FeatureTypeStyle} represents the style that converts an instance of \texttt{MapContentType} to visualizations (graphics). They are associated through the relation \texttt{hasStyle}. Following the design pattern of the knowledge base for geovisualization from \cite{huang2020towards}, we design the cartography module coupling ontology and semantic rules.

For the ontological part, the concept \texttt{FeatureTypeStyle} is associated with \texttt{Symbol}, and thereafter \texttt{Symbolizer} concept, which then is linked to specific geometry portrayal concepts, e.g.,\texttt{Stroke} for linestrings, and \texttt{Fill} for polygons. \texttt{Symbol} is also associated with \texttt{Legend} and \texttt{LegendItem} which represent the information of the map legend. An instance of \texttt{Symbol} could have a number of instances of \texttt{Symbolizer}, which is at the implementation level to link particular portrayal rules with specific geometry portrayal concepts. For instance, we can render the battles lasting longer than a year (rule condition) with a particular size and color of dots (rule conclusion - a particular symbolizer). 

The portrayal rules in this paper are organized in a rule base, and implemented with SPARQL rules. Here we provide four concrete examples of such rules in Listings \ref{q:us_part} - \ref{q:start_time}. The first two rules are based on the relation among entities. The rest two rules are based on temporal constraints.
Listing \ref{q:us_part} is a SPARQL rule that states using a particular symbolizer for the battles during the World War II that the US participated. The condition of the rule comes after the keyword \texttt{WHERE}, saying that: 1) the entity is a battle (the first clause in the condition); 2) the battle is a part of World War II (the second and third clause); and 3) the US participated in the battle (the fourth clause). Note that all the predicates (e.g., wdt:P32) and entities (e.g., wd:Q362) are from Wikidata. The rule conclusion comes above after the keyword \texttt{CONSTRUCT}, saying that this rule uses a particular symbolizer (symbolizer\_0 in this case) for the entities that meet the conditions below. Likewise, Listing \ref{q:num_part}, \ref{q:duration}, and \ref{q:start_time} formalize the rules \textit{using a particular symbolizer for the battles with more than 5 participating countries}, \textit{using a particular symbolizer for the battles lasting more than 30 days}, as well as \textit{using a particular symbolizer for the battles whose start time is in 1939}. 
Listing \ref{q:duration} and \ref{q:start_time} show SPARQL rules with temporal constraints which are particularly interesting from the narrative cartography perspective.

Note that here we show how to visualize a specific set of battles that satisfy certain conditions with the Wikidata as the underlining map content KG. As for the map content KG built based on the \textit{Map Content} ontology design pattern, we need to change the selection clauses to select \texttt{MapContentItem} instances that satisfy the said condition. As for the rules depicted in Listing \ref{q:duration} and \ref{q:start_time}, the start time of battles can be access through the \texttt{TemporalExtent} of each \texttt{MapContentItem} instance. The reason we use Wikidata as the map content KG is that it is easier for the reader to test these rules through real SPARQL queries shown in each listing.

In terms of the implementation of the portrayal rule base, such rules can be encapsulated in a named graph, and represented with the SHACL rule vocabulary\footnote{\url{https://www.w3.org/TR/shacl/}}. In some RDF (KG) stores/frameworks (e.g., RDF4J\footnote{\url{https://rdf4j.org/}}), deductions can be derived automatically, and in such settings the rule base could derive corresponding symbolizers to individual map objects to realize a knowledge-based narrative cartography scenario. For technical details, please see \cite{huang2020towards} and its Github repository\footnote{\url{https://github.com/RightBank/Knowledge-based-geovisualisation}}.

By modeling the whole geovisualization process with our \textit{Cartography} ontology design pattern, we can explicitly express the semantics behind each map symbol and legend item. For example, when we see a specific map symbol \texttt{portrayal:symbolizer\_2} (see Listing \ref{q:duration}) on a narrative map, it indicates a battle lasting more than 30 days during the World War II. This practice can help us to overcome the semantic challenges in geovisualization described in Section \ref{sec:intro} and \ref{sec:geoenrichment_limit}, and achieve map reproducibility.

\begin{figure}[h]
	\centering
	\includegraphics[width=1.0\textwidth]{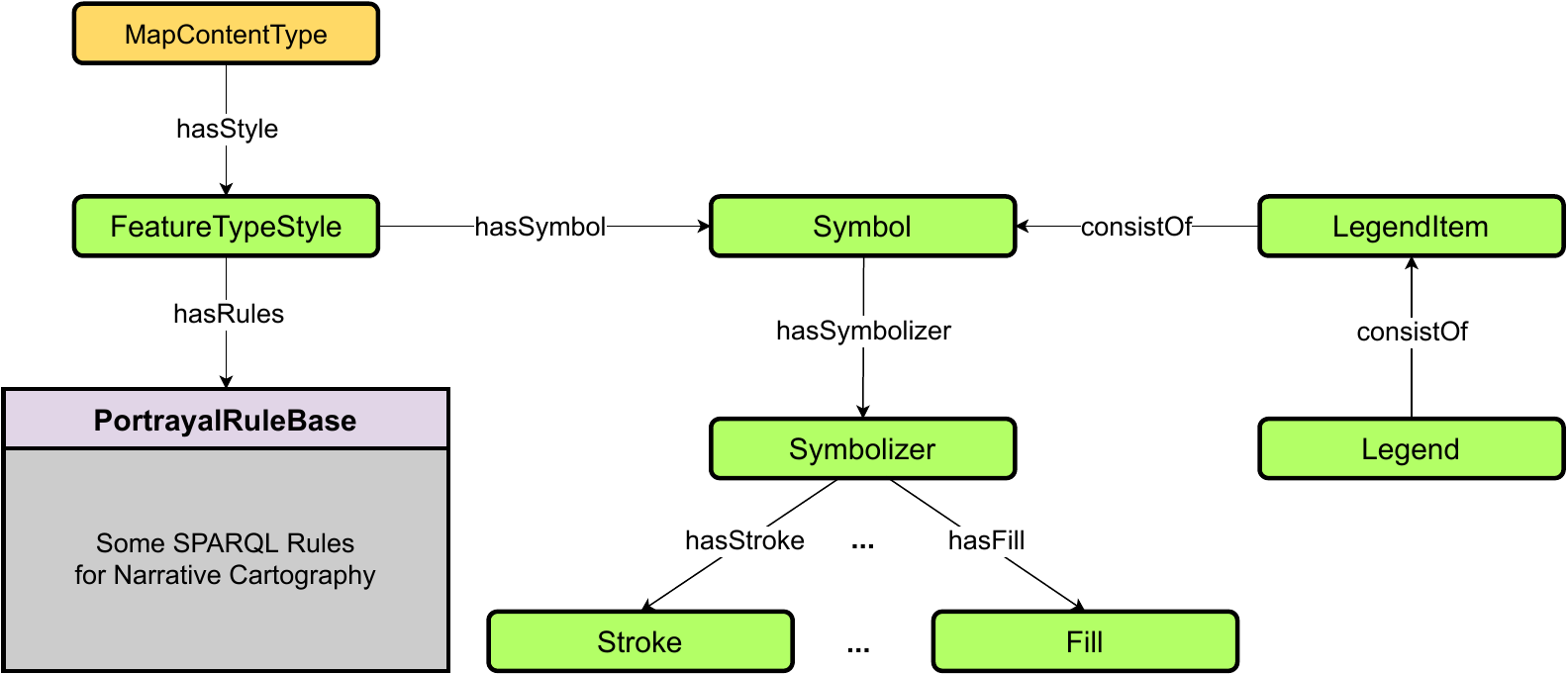}
	\caption{The \textit{Cartography} module of the \textit{Narrative Cartography Ontology}.}
	\label{fig:carto-odp}
\end{figure}

\vspace{0.5cm}
\hspace{-0.5cm}
\begin{minipage}[c]{\columnwidth}
	\begin{lstlisting}[
	linewidth=\columnwidth,
	breaklines=true,
	basicstyle=\ttfamily, 
	captionpos=b, 
	caption={Use a particular symbolizer for the battles during World War II that the US participated. See Link\textsuperscript{a} for it corresponding SPARQL query in Wikidata.
	\small\textsuperscript{a} \url{https://api.triplydb.com/s/NDF05YYBl}
	}, 
	label={q:us_part},
	frame=single
	]
@[prefix definitions]
CONSTRUCT {?battle symbolizer:isSymbolizedBy portrayal:symbolizer_0}
WHERE{
     ?battle wdt:P31 wd:Q178561.       # the entity is a battle
     {wd:Q362 wdt:P527* ?battle .}     # the entity is x-degree away from the World War II node with ``has part'' relation  
     union
     {?battle wdt:P361* wd:Q362 .}     # the entity is x-degree away from the World War II node with reverse ``part of'' relation  
     ?battle wdt:P710 wd:Q30.          # retrieve the battle that the US was involved in
}
	\end{lstlisting}
\end{minipage}

\hspace{-0.5cm}
\begin{minipage}[c]{\columnwidth}
	\begin{lstlisting}[
	linewidth=\columnwidth,
	breaklines=true,
	basicstyle=\ttfamily, 
	captionpos=b, 
	caption={Use a particular symbolizer for the battles during World War II with more than 5 participating countries. See Link\textsuperscript{a} for it corresponding SPARQL query in Wikidata.
	\small\textsuperscript{a} \url{https://api.triplydb.com/s/tY6amm4AZ}
	}, 
	label={q:num_part},
	frame=single
	]
@[prefix definitions]
CONSTRUCT {?battle symbolizer:isSymbolizedBy portrayal:symbolizer_1}
WHERE
{
  Select ?battle (COUNT(?participant_country) AS ?number_of_participants)
  WHERE{
    ?battle wdt:P31 wd:Q178561.            # the entity is a battle
    {wd:Q362 wdt:P527* ?battle .}          # the entity is x-degree away from the World War II node with ``has part'' relation  
    union
    {?battle wdt:P361* wd:Q362 .}          # the entity is x-degree away from the World War II node with reverse ``part of'' relation  
    ?battle wdt:P710 ?participant_country. # retrieve the participating countries
  }
  GROUP BY ?battle
  HAVING (?number_of_participants > 5)     # filter out the battles having more than 5 participating countries
}
	\end{lstlisting}
\end{minipage}

\hspace{-0.5cm}
\begin{minipage}[c]{\columnwidth}
	\begin{lstlisting}[
	linewidth=\columnwidth,
	breaklines=true,
	basicstyle=\ttfamily, 
	captionpos=b, 
	caption={Use a particular symbolizer for the battles during World War II which lasted more than 30 days. See Link\textsuperscript{a} for it corresponding SPARQL query in Wikidata.
	\small\textsuperscript{a} \url{https://api.triplydb.com/s/sUOFyqZNx}
	}, 
	label={q:duration},
	frame=single
	]
@[prefix definitions]
CONSTRUCT {?battle symbolizer:isSymbolizedBy portrayal:symbolizer_2}
WHERE{
    ?battle  wdt:P31 wd:Q178561.      # the entity is a battle
    {wd:Q362 wdt:P527* ?battle .}     # the entity is x-degree away from the World War II node with ``has part'' relation  
    union
    {?battle wdt:P361* wd:Q362 .}     # the entity is x-degree away from the World War II node with reverse ``part of'' relation  
    ?battle  wdt:P580 ?start_time.    # start time of the battle
    ?battle  wdt:P582 ?end_time.      # end time of the battle
    bind((?end_time - ?start_time) as ?duration ) # derive battle duration
    FILTER (?duration > 30)           # filter out the battles longer than 30 days.
}
	\end{lstlisting}
\end{minipage}

\hspace{-0.5cm}
\begin{minipage}{\columnwidth}
	\begin{lstlisting}[
	linewidth=\columnwidth,
	breaklines=true,
	basicstyle=\ttfamily, 
	captionpos=b, 
	caption={Use a particular symbolizer for the battles during World War II whose start time is in 1939. See Link\textsuperscript{a} for it corresponding SPARQL query in Wikidata.
	\small\textsuperscript{a} \url{https://api.triplydb.com/s/ETk6mbUHx}
	}, 
	label={q:start_time},
	frame=single
	]
@[prefix definitions]
CONSTRUCT {?battle symbolizer:isSymbolizedBy portrayal:symbolizer_3}
WHERE
{
  ?battle wdt:P31 wd:Q178561.            # the entity is a battle
  {wd:Q362 wdt:P527* ?battle .}          # the entity is x-degree away from the World War II node with ``has part'' relation  
  union
  {?battle wdt:P361* wd:Q362 .}          # the entity is x-degree away from the World War II node with reverse ``part of'' relation  
  ?battle wdt:P580 ?start_time .         # Query the battle's start time
  FILTER(?start_time > "1939-01-01T00:00:000"^^xsd:dateTime)  # Keep the battles whose start time is later than Jan 1, 1939
  FILTER(?start_time < "1940-01-01T00:00:000"^^xsd:dateTime)  # Keep the battles whose start time is earlier than Jan 1, 1940    
}
order by ?start_time
\end{lstlisting}
\end{minipage}

 \section{Conclusions and Outlook}   \label{sec:conclude}

In this work, we introduce the idea of doing narrative cartography with knowledge graphs. The main motivation is to overcome 
\reviseone{the data acquisition \& integration challenge and the semantic challenge}
of the conventional narrative cartography techniques and foster underlying data integration, data reusablity, and visualization reproducibility. 
We first discuss a way to utilize our KG-based GeoEnrichment tool developed for ArcGIS Pro to directly make narrative maps based on an existing KG - Wikidata. Two use cases are provided to illustrate the effectiveness of this idea - a map of Ferdinand Magellan's expedition as well as a map of all events during World War II. We show that this KG-based GeoEnrichment tool can effectively help map a narrative with substantially less efforts
\reviseone{in data acquisition, preprocessing, and integration. Moreover, }
our approach requires nearly no prior knowledge about Semantic Web technologies from the users. 

\reviseone{We} also identify several limitations for this GeoEnrichment approach - data incompleteness of the existing KGs, semantic incompatibility \reviseone{in map data among different data sources}, as well as the semantic challenges of geovisualization process. To overcome the last two challenges, we develop a modular ontology for narrative cartography which consists of two ontology design patterns - \textit{Map Content} module and \textit{Cartography} module. The \textit{Map Content} module formalizes the concepts and relations entailed in the map content. It can be utilized to formally define the semantics of each map content concept and achieve semantic interoperability between the narrative map KG and other existing KGs. So the semantic incompatibility issue can be well handled. The \textit{Cartography} module explicitly models the semantics behind the geovisualization process so that a narrative map can be easily reproduced through the deductions made by portrayal rules.

\reviseone{This paper can be treated as the latest endeavor to use KGs and Semantic Web technologies for a narrative cartography purpose. Compared with previous work discussed in Section \ref{sec:relwork-carto} that mostly focused on modeling map content for map sharing and search purpose, our narrative cartography ontology formalizes both the map content as well as the geovisualization process. }
We have discussed many advantages of this approach, including a semantic explicit map data representation to facilitate map data reusability, a more expressive way to represent the geovisualization process to facilitate map reproducibility, and an easier way to do data acquisition and integration. 

However, in order to establish a mature KG-based narrative cartography framework, there are still several technical challenges to be resolved. First, as discussed in Section \ref{sec:geoenrichment_limit}, the data incompleteness issue of the existing KGs can possibly affect the quality of the map data which in turn affects the reliability of the resulting maps. Recently we have witnessed many efforts to solve this data incompleteness issue by using relational machine learning techniques~\citep{dong2014knowledge,nickel2015review}. Various KG embedding techniques have been proposed for link prediction and KG completion, such as TransE~\citep{bordes2013translating}, TransH~\citep{wang2014knowledge}, TransR~\citep{lin2015learning}, ComplEx~\citep{trouillon2016complex}, R-GCN~\citep{schlichtkrull2018modeling},
TransGCN~\citep{cai2019transgcn}, RoteE~\citep{sun2019rotate}, and so on. They perform very well on some experimental datasets like FB15K and WN18. However, how well they perform in a real-world setting and how reliable the predicted links are have not been systematically studied. Moreover, most KGE models ignore literal nodes while focusing on predicting relations among entities. These literal nodes are particularly important for geovisualization purpose such as geographic coordinates, timestamps, and other text information. Some recent works have shown how to encode spatial information \citep{mai2020se,mai2021review} and temporal information \citep{dasgupta2018hyte,kazemi2019time2vec,cai2021time} into the embedding space so that link prediction among entities and certain literal nodes (geographic coordinates, timestamps) are possible. Yet, they also need to be validated for their reliability.

Second, although our proposed \reviseone{\textit{Linked Data Relationship Finder}} tool (in Figure \ref{fig:relfinder-exp-start},  \ref{fig:relfinder-exp-via}, and \ref{fig:relfinder-ww2}) is very useful to explore N-degree relation paths from some start nodes, it also has some limitations. For example, it can not fully support a logical disjunction among several relationship paths such as Listing \ref{q:ww2_subevent} which is sometimes important for a narrative cartography use case. This is due to the user interface design of this tool and the restriction of ArcGIS Python toolbox. To design a more intelligent user interface for GIS users to interact with KGs from within GISystems, we need to do more study on the user need and have a more comprehensive list of competency questions.

Third, the proposed modular ontology for narrative cartography only focuses on the main body of the map - the map itself and the legend. We have not discussed how to represent multimedia data such as images, audios, and videos in the \textit{Map Content} module. We also have not discussed how to display these information during the geovisualization process through the \textit{Cartography} module. We treat this as the future work, whereas a preliminary prototype of developing geovisualizations entirely backed by KGs can be found in \cite{huang2020towards}.

Fourth, through the KG-based GeoEnrichment tool, we have shown how to use an existing KG as the map content KG for narrative mapping. However, we have not discussed how to use the proposed \textit{Cartography} module to guide the mapping process within an existing GISystem.

Last but not least, in this work, we focus on representing the real-world historical events and objects mentioned in narratives into a KG and visualizing them on a map. This work does not cover how to model and represent the content of fictional narratives. Recently, \cite{branch2017representing} have made progresses in formalizing transmedia fictional world into an ontology which describes the relations among different fictional concepts such as characters, elements of power, items, places, events, and so on. How to develop geovisualization on top of them is an interesting yet challenging task. For instance, since the fictional world might describe a totally different geographic space and layout, commonly used basemaps used for geovisualization may not be applicable here. Making fictional map is nontrivial task. We leave this as the future work.

\section*{Acknowledgements}

We would like to acknowledge the guidance and suggestion of Prof. Krzysztof Janowicz, and Prof. Pascal Hitzler on the ontology design pattern development. 
We would like to thank the three anonymous reviewers for their thoughtful comments and suggestions. 

This work is mainly funded by the National Science Foundation under Grant No. 2033521 A1 -- KnowWhereGraph: Enriching and Linking Cross-Domain Knowledge Graphs using Spatially-Explicit AI Technologies. Gengchen Mai acknowledges the support by the Office of the Director of National Intelligence (ODNI), Intelligence Advanced Research Projects Activity (IARPA), via 2021-2011000004. Weiming Huang acknowledges the support by open access funding provided by Lund University and the National Natural Science Foundation of China (Grant Number: 42101421). 
Any views, opinions, findings, and conclusions or recommendations expressed in this material are those of the authors and should not be interpreted as necessarily representing the official policies, either expressed or implied, of NSF, ODNI, IARPA, or the U.S. Government. The U.S. Government is authorized to reproduce and distribute reprints for governmental purposes not-withstanding any copyright annotation therein.

\bibliographystyle{spbasic}     

\end{document}